\documentclass{article}
\usepackage{geometry}
\usepackage{graphicx} 
\usepackage{epsfig}
\usepackage{subcaption} 
\usepackage{times}
\usepackage{epsfig}
\usepackage{subcaption}
\usepackage{graphicx}
\usepackage{amsmath}
\usepackage{amssymb}

\usepackage[rightcaption]{sidecap}
\usepackage{algorithmic}
\usepackage{amsopn}
\usepackage{tabularx, booktabs}
\usepackage{mathtools}
\usepackage[colorinlistoftodos]{todonotes}
\usepackage[ruled,vlined,linesnumbered,lined,boxed,commentsnumbered]{algorithm2e}

\usepackage{tabularx, booktabs}
\usepackage{subfig}
\usepackage[symbol]{footmisc}
\usepackage{latexsym}
\usepackage{comment}
\newtheorem{Lemma}{Lemma}

\newtheorem{Theorem}{Theorem}

\usepackage{xcolor}
\usepackage{wrapfig}
\usepackage{comment}
\newtheorem{definition}{Definition}
\newtheorem{Corollary}{Corollary}

\title{MS-IMAP - A Multi-Scale Graph Embedding Approach for Interpretable Manifold Learning}
\author{Shay Deutsch \footnote{Corresponding author.}  \footnotemark[4]{}  \\
{\tt\small shaydeu@gmail.com}
\and
Lionel Yelibi \footnotemark[4] \\
{\tt\small ylblio001@myuct.ac.za}
\and 
Alex Tong Lin  \footnotemark[4] \\
{\tt\small atlin271@gmail.com}
\and 
Arjun Ravi Kannan \footnotemark[4]\\
{\tt\small arjun.kannan@gmail.com}
  }
\date{}

\begin{document}

\maketitle

\begin{abstract}
Deriving meaningful representations from complex, high-dimensional data in unsupervised settings is crucial across diverse machine learning applications. This paper introduces a framework for multi-scale graph network embedding based on spectral graph wavelets that employs a contrastive learning approach. We theoretically show that in Paley-Wiener spaces on combinatorial graphs, the spectral graph wavelets operator provides greater flexibility and control over smoothness compared to the Laplacian operator, motivating our approach. A key advantage of the proposed embedding is its ability to establish a correspondence between the embedding and input feature spaces, enabling the derivation of feature importance. We validate the effectiveness of our graph embedding framework on multiple public datasets across various downstream tasks, including clustering and unsupervised feature importance.
\end{abstract}

\section{Introduction}
\label{submission}

Graph Embeddings and Manifold Learning (\cite{LLE, Belkin:2003, Tenenbaum00, t-SNE, Coifman05}) play a pivotal role in analyzing complex data structures encountered in a wide range of machine learning applications. The representations learned by these techniques uncover the intrinsic geometry of the data and support downstream tasks such as clustering and visualization—especially in scenarios where labels are unavailable or unreliable.
Most manifold learning methods typically perform nonlinear dimensionality reduction to embed data into lower-dimensional spaces, while some also explore high-dimensional embeddings (\cite{Scattering_Graph, HKS}). Nonetheless, most such methods lack a direct connection to the input features, preventing the evaluation of individual feature contributions and resulting in embeddings with limited interpretability. This limitation is critical in domains like finance, where key behavioral indicators must be traced back to individual features, or in biology, where interpretability is vital for uncovering mechanisms within complex systems. To address these limitations, this paper makes three primary contributions:

1. We introduce a framework that leverages multi-scale graph representation using a 3D tensor and contrastive learning. This enhances the expressiveness of the embedding by jointly optimizing over both low- and high-frequency components, enabling the capture of fine-grained data structure.

2. We propose a manifold-based interpretation method for feature importance, allowing each dimension of the learned embedding to be directly linked back to a unique original feature.

3. We characterize the theoretical representational capacity of the Spectral Graph Wavelet (SGW) operator by analyzing functions from Paley-Wiener spaces (\cite{Pesenson2008SamplingIP}) on combinatorial graphs. This analysis demonstrates that SGW provides greater flexibility and smoother control compared to traditional Laplacian-based operators.

Our method, MS-IMAP, is designed to associate each embedding coordinate with a unique input feature, thereby enabling the estimation of feature importance relative to the learned embedding space. This interpretability arises from a deliberate design choice: each embedding coordinate corresponds to a single, unique input feature and is optimized through spectral graph wavelets to align with the underlying graph structure. Unlike standard manifold learning and graph embedding methods, where each embedding dimension typically blends information from all input features, MS-IMAP preserves the semantic identity of each feature by assigning it to a distinct embedding dimension.

Another key distinction of our approach—compared to both classical (\cite{Belkin:2003}) and modern techniques (\cite{UMAP}), which typically rely on the low-frequency components of the graph Laplacian and thus capture only coarse structural information—is the use of multi-scale graph representations via Spectral Graph Wavelets (SGW) within a contrastive learning framework. This allows our method to retain or even expand the dimensionality of the data, yielding richer and more expressive representations while preserving interpretability. We further employ stochastic gradient descent (SGD) with a novel 3D-tensor-based contrastive objective to optimize embeddings that capture both global and local graph structure. We empirically demonstrate that applying feature importance techniques to MS-IMAP embeddings yields more informative feature subsets for clustering than applying them to raw input data. This provides strong evidence of semantic alignment—often absent in nonlinear methods such as deep networks or conventional manifold learning. MS-IMAP achieves interpretable and semantically faithful embeddings, while remaining competitive with state-of-the-art graph embedding techniques across a range of tasks.

\begin{figure*}[t]
    \centering
    \includegraphics[width=0.49\linewidth]{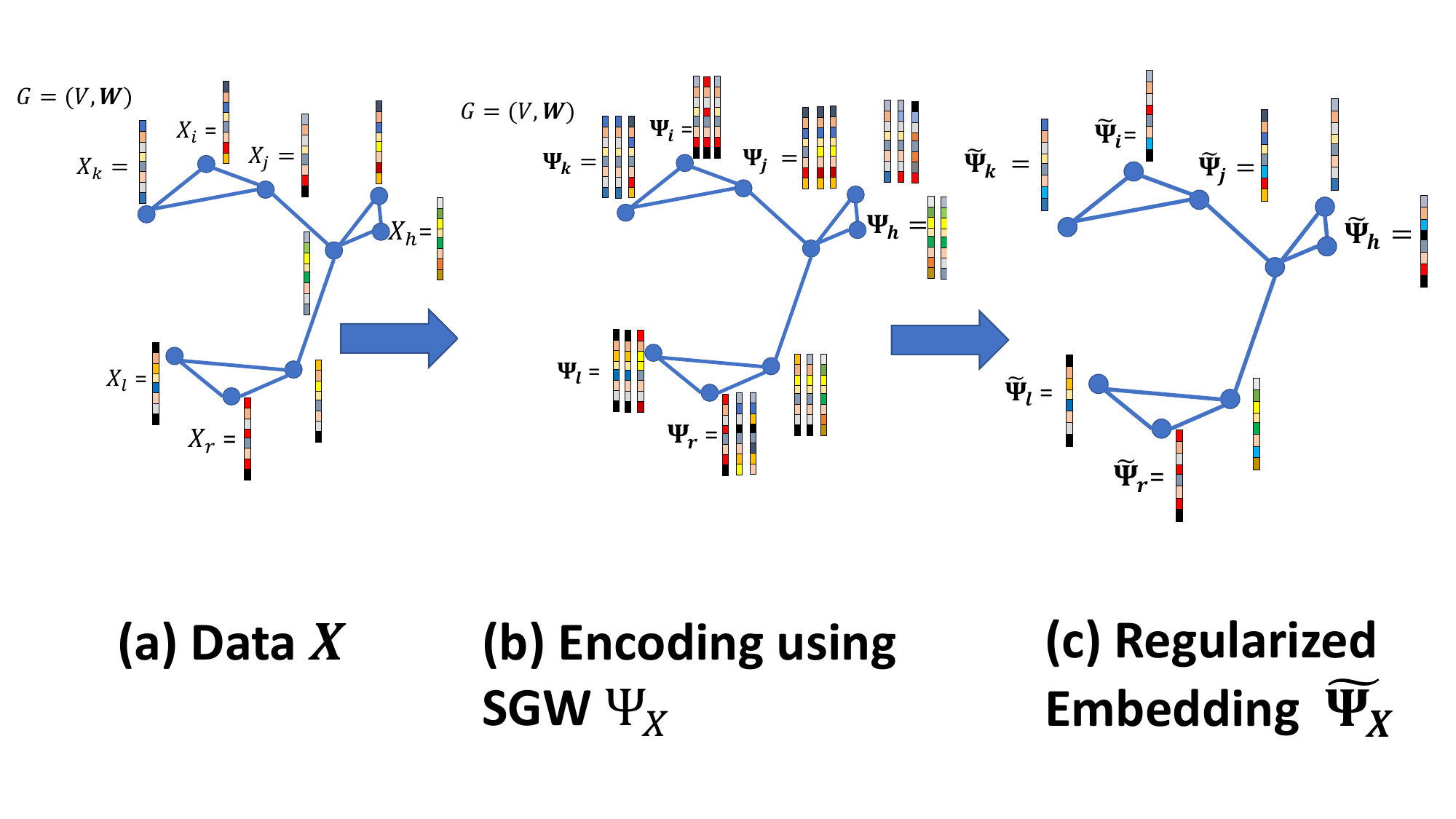}
    \hfill
    \includegraphics[width=0.4\linewidth]{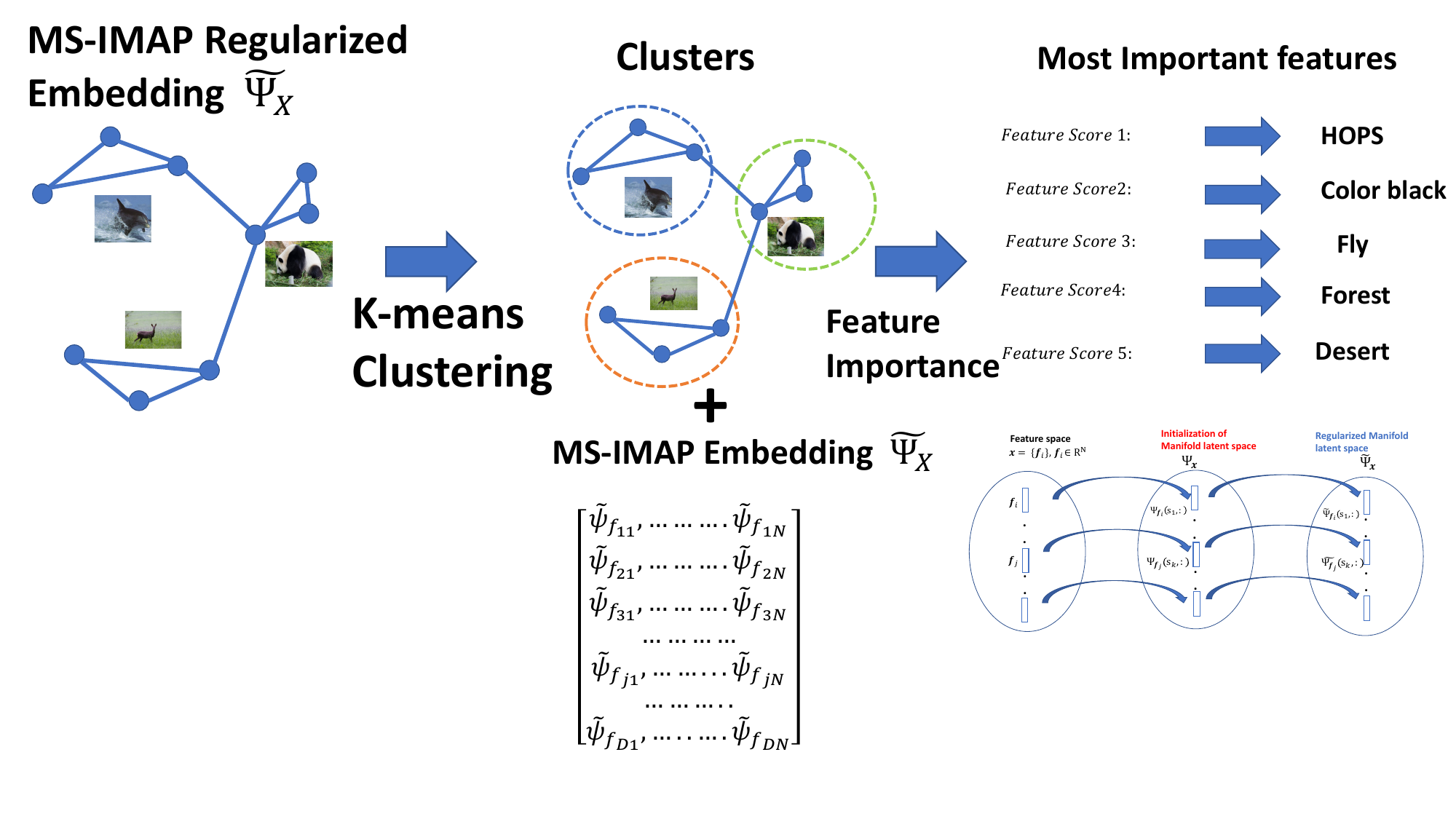}
    \caption{Illustration of the proposed method. (Left) Initial encoding using SGW in (b) and regularized embedding in (c). (Right) Feature embedding space representation, which has a one-to-one correspondence with the input features, can be used along with clustering for interpretability models, such as feature importance.}
    \label{fig:combined_illustration}
\end{figure*}

\section{Related Work}
\label{RE_W}

Extensive efforts have been dedicated to exploring manifold learning methods that aim to produce low-dimensional embeddings while preserving the intrinsic structure of high-dimensional data. Classical manifold learning techniques include Locally Linear Embedding (LLE) \cite{LLE}, Laplacian Eigenmaps (\cite{Belkin:2003}), Isomap (\cite{Tenenbaum00}), Diffusion Maps (\cite{CoifmanDM}), and t-distributed Stochastic Neighbor Embedding (t-SNE) (\cite{t-SNE}).

One of the central challenges in these methods is maintaining a balance between local and global structure, especially in the presence of noise. To address this, various denoising strategies have been developed (\cite{MDhein, MFD, ZSLwavelets, DeutschM17, ZSL_Isoperimteric, DeutschOM18, ijcai15, SSVM15}), aiming to improve robustness and extend the applicability of these techniques to real-world, noisy datasets.

UMAP (\cite{UMAP}) has shown strong empirical performance for visualization and cluster separation. It typically employs Laplacian Eigenmaps initialization, which emphasizes low-frequency graph signals while neglecting higher-frequency patterns that may carry important information. 

Recent work (\cite{UMAP_True_Loss}) has shown that negative and positive sampling in UMAP significantly influences the effective loss function. These insights have led to studies on alternative sampling schemes (\cite{NEURIPS2023_15dc2344}), bounds on generalization error in embedding methods (\cite{NEURIPS2021_09779bb7}), and biases introduced by high-degree nodes during graph construction and sampling (\cite{NEURIPS2021_ca954182}).

Manifold learning and non-linear dimensionality reduction serve purposes beyond visualization; one of their key goals is to learn representations that capture intrinsic geometric structure and approximate geodesic distances. The learned embedding space dimensionality can go beyond two or three dimensions. However, increasing dimensionality does not necessarily improve robustness: noise and the curse of dimensionality can introduce instability. These methods also suffer from computational limitations. Moreover, to compute embedding in higher dimensions many algorithms—such as Laplacian Eigenmaps and UMAP—require computing large portions of the Laplacian spectrum, making them computationally intensive for large datasets. Our method addresses these issues by producing interpretable and efficient embeddings while leveraging spectral information across multiple frequency scales.

A key limitation of many manifold learning and graph embedding methods is the absence of an explicit and interpretable mapping, which prevents evaluating the contribution of each input feature to the resulting low-dimensional embedding. This gap restricts their utility in domains where interpretability is crucial. Providing such a link can enable improved feature selection, clearer interpretation of embeddings, and greater robustness to noise (\cite{10.1007/978-3-030-46150-8_19, frye2021shapley, pmlr-v151-zharmagambetov22a}).

Spectral Graph Wavelets (SGW) are central to our method (see Section \ref{sec:Feature_Representation}), offering localization in both spectral and vertex domains. SGW apply a family of bandpass filters to the graph Laplacian's spectrum, allowing for multi-resolution analysis of graph signals and providing richer representations across frequency bands. Our approach is also related to graph embedding methods that exploit multiscale structure, such as Diffusion Wavelets (\cite{DW}), Geometric Scattering for Graph Data Analysis (\cite{Gao2018GeometricSF}), and the Graph Scattering Transform (\cite{Scattering_Graph}), which extract graph features for downstream tasks like classification. These methods often embed data in higher-dimensional spaces to improve representation power. 

\subsection{Preliminaries}

Consider a set of points $\mathbf{x}=\left \{ \mathbf{x}_{i} \right  \}, i=1,...N, \mathbf{x}_{i} \in \mathbb{R}^{D}$, where each $ \mathbf{x}_{i} \in \mathbb{R}^D$, sampled from an unknown smooth manifold $M \subset \mathbb{R}^D$. To model the geometric structure of the data, we construct an undirected, weighted graph $G = (V, \mathbf{W})$ over the point set $\mathbf{x}$, where $V$ is the set of nodes and $\mathbf{W} = (w_{ij})$ is the weighted adjacency matrix encoding edge weights between nodes. The weights $w_{ij}$ can be computed using various techniques. A common approach is to apply a Gaussian kernel to pairs of points $ \mathbf{x}_{i}$ and $\mathbf{x}_j$ within the $k$ nearest neighbors of $ \mathbf{x}_{i}$, denoted as $\text{kNN}( \mathbf{x}_{i})$. In this work, we adopt the adaptive Gaussian kernel construction proposed in \cite{UMAP}, though other graph construction methods may also be employed. Let $\mathbf{L}$ denote the combinatorial graph Laplacian defined as $\mathbf{L} = \mathbf{D} - \mathbf{W}$, where $\mathbf{D}$ is the diagonal degree matrix with entries $d_{ii} = d(i)$, and the degree $d(i)$ of vertex $i$ is the sum of the weights of all edges incident to $i$. We use the normalized graph Laplacian, defined as: $$\mathbf{L}_N = \mathbf{D}^{-1/2} \mathbf{L} \mathbf{D}^{-1/2} = \mathbf{I} - \mathbf{D}^{-1/2} \mathbf{W} \mathbf{D}^{-1/2}$$ which has real eigenvalues in the interval $[0, 2]$. For simplicity, we refer to $\mathbf{L}_N$ as $\mathbf{L}$ for the remainder of the paper. Let $\lambda_1, \ldots, \lambda_N$ and $\boldsymbol{\phi}_1, \ldots, \boldsymbol{\phi}_N$ denote the eigenvalues and corresponding eigenvectors of $\mathbf{L}$. We denote the matrix of eigenvectors as $\Phi \in \mathbb{R}^{N \times N}$ and the diagonal matrix of eigenvalues as $\Lambda = \text{diag}(\lambda_1, \ldots, \lambda_N)$. In our setting, each input coordinate $f_r \in \mathbb{R}^N$ — viewed as a graph signal—is defined across the $N$ nodes of the graph. The original data matrix $\mathbf{x}$ can thus be expressed as a collection of such signals:  $\mathbf{x} = \left (f_{1}, f_{2},.. ,f_{r},.. f_{D} \right )$. Our goal is to derive a multi-scale embedding from these signals that approximates the intrinsic manifold coordinates, using spectral information derived from the graph Laplacian.

\subsection{Multi-scale representations using SGW} 

In the past two decades, several multi-scale representations for data on irregular graphs have been developed, including Spectral Graph Wavelets (SGW) \cite{Hammond} and Diffusion Wavelets \cite{DW}. In this work, we focus on the SGW-based multi-scale graph transform. SGW offers a principled way to balance spectral and spatial resolution, as its coefficients are localized in both domains. These wavelets are constructed using a kernel function operator \( g(\mathbf{L}) \), which acts on a signal by modulating each of its spectral components (i.e., Fourier modes) \cite{Hammond}. This design enables a trade-off between vertex-domain (spatial) and frequency-domain (spectral) localization. The spatial localization is implicitly controlled by a single scale parameter defined in the spectral domain: greater localization in the vertex domain corresponds to a broader spectral bandwidth. To represent a signal \( f \in \mathbb{R}^{N} \) at multiple scales \( S = [s_1, s_2, \ldots, s_K] \), the SGW transform is defined as follows. Let \( g(\lambda) \) be a bandpass filter in the spectral domain. Let \( \delta_i \in \mathbb{R}^{N} \) denote the delta function centered at vertex \( i \in G \), where \( \delta_i(j) = 1 \) if \( i = j \), and \( \delta_i(j) = 0 \) otherwise. A wavelet centered at node \( i \) and scale \( s \) is then given by: $$\psi(s, i) = \Phi g(s \lambda) \Phi^{T} \delta_i $$.The value of \( \psi(s, i) \) at vertex \( m \) can be written explicitly as: $$\psi(s, i)(m) = \sum_{l=1}^{N} g(s \lambda_l) \phi_l^*(i) \phi_l(m)$$. Given a graph signal \( f \in \mathbb{R}^N \), the \textit{\textbf{SGW coefficient}} at node \( i \) and scale \( s \) is defined as:

\begin{equation}
    \psi_f(s, i) = \sum_{l=1}^{N} g(s \lambda_l) \hat{f}(\lambda_l) \phi_l(i)
\end{equation}

\textbf{Fast computation using Chebyshev polynomials:}  \\
Direct computation of SGW coefficients is computationally expensive, requiring \( O(N^3) \) operations for \( N \) nodes. To address this, Hammond et al. \cite{Hammond} proposed an efficient algorithm based on approximating the scaled generating kernels \( g(s \lambda) \) using low-order Chebyshev polynomials of the Laplacian \( \mathbf{L} \), which are applied directly to the input signal. This significantly reduces computational cost (see Appendix for further details).
\section{Our Proposed framework: Multi-Scale IMAP}
\label{sec:Proposed_framework}
In this section, we introduce Multi-scale IMAP, a framework for interpretable embedding via manifold learning utilizing a multi-scale graph representation. This approach enables us to maintain global regularity and preserve local structure without sacrificing interpretability. Our method imposes a differentiable structure on the mapping $h: \mathbf{x} \rightarrow \psi_{\mathbf{x}}$, supported on a discrete graph $G = (V,\mathbf{W})$, where $\psi_{\mathbf{x}}$ represents the encoded multi-scale graph transform of $ \mathbf{x}$. Multi-scale IMAP consists of two main steps:  In \textbf{Step 1}, our approach constructs a multi-scale graph representation by incorporating feature signals through the SGW Transform across multiple scales and graph frequencies. We suggest a 3D tensor-based approach (detailed in Algorithm 1), where we construct a 3D tensor with dimensions ${K} \times {N}\times{D}$ to provide an effective representation for encoding multi-scale transform. We leverage the encoded 3D tensor structure to align the optimized embedding with all scales simultaneously to enforce differentiable structure of the transformed features. 
The 3D tensor-based approach (detailed in Algorithm 1) provides the most effective representation for the following reasons: \\
(i) Joint and simultaneous optimization: using the 3D tensor-based method enables the joint and simultaneous optimization of all scales for each sample by leveraging the SGW coefficients of each spectral band within the cross-entropy loss framework while adhering to the fixed graph structure. This approach effectively balances coarse and fine-grain resolutions, with summation performed only after the contrastive learning optimization process is complete. By smoothing both low- and high-frequency information jointly and simultaneously with respect to the graph structure, this method reduces the reliance on threshold-based techniques for removing high-frequency spectral bands, which may be less efficient. Jointly optimizing across scales ensures consistency in the optimization process for all scales with respect to a fixed pair of nodes. Although there may be correlations between overlapping SGWs corresponding to different spectral bands, each scale is associated with a distinct bandwidth of frequencies, effectively capturing graph structures at varying levels of neighborhood information. By integrating multiple scales, SGWs capture neighborhood information continuously across varying granularities, enabling a representation that strikes a balance between local and global structural details.  \\
(ii) Enhanced Interpretability: the proposed representation establishes a one-to-one correspondence between input features and the final embedding features, making feature interpretation more straightforward and intuitive. This is demonstrated in Section \ref{sec:Interpretable Embedding}, where we show that the embeddings produced enable effective feature importance rankings. By ensuring this direct correspondence, our approach facilitates meaningful interpretation of the embedding space. This combination of optimization efficiency and enhanced interpretability highlights the clear advantages of the 3D tensor-based representation. We note that an alternative to the 3D tensor-based representation approach is to use simple concatenation, where all features are transformed using SGW and each data point is represented by a vector formed by concatenating all corresponding SGW coefficients.  For completeness we describe this alternative in the Appendix. In \textbf{Step 2}, We integrate the multi-scale representation within the SGD optimization framework, simultaneously leveraging both low and high-frequency information. This integration leads to fine-grained manifold regularization and improved robustness in downstream tasks.

\subsection{Step 1: Feature Representation to encode multi-scale structure}
\label{sec:Feature_Representation}
To encode multi-scale representations used for subsequent optimization we focus on a 3D tensor based method. Note that each dimension in the embedding space is constructed using a single feature in the original feature set, which is an essential characteristic that can be leveraged for interpreting graph embeddings. \\
\textbf{3D-Tensor Encoding}: The 3D-Tensor Encoding method concatenates the multi-scale representation for all features at a fixed scale, generating a matrix representation $\psi_{\mathbf{x}}(s_{j},:,:) \in\mathbb{R}^{D}\times\mathbb{R}^{N}$ for each scale $s_{j}$, encoding SGW coefficients across features and nodes. After concatenating all $f_{i}$ at a fixed scale $s_j$, we have
$$\psi_{\mathbf{x}}(s_{j},:,:) = \psi_{{f}_{1}}(s_j,:) \| \psi_{{f}_{2}}(s_j,:) \| ... \| \psi_{{f}_{D}}(s_j,:)$$
where we designate the concatenation using $\left |  \right |$, with  $\mathbf{c} ( \psi_{{f}_{i}}(s_k,:,:),  \psi_{ {f}_{j}}(s_k,:,:) ) = \psi_{ {f}_{i}}(s_k,:,:) \left |  \right |  \psi_{ {f}_{j}}(s_k,:,:)$. The optimization can be performed separately for each scale $s_{j}$, or jointly for all scales $s_{j}$ using the 3D tensor  $\psi_{\mathbf{x}} \in \mathbb{R}^{K} \times\mathbb{R}^{D}\times\mathbb{R}^{N}$ . We detail the proposed feature representation encoding methods in the pseudo code algorithm \ref{alg:Encoding_multi-scale}.

\begin{algorithm}[tb]
   \caption{Encoding multi-scale structure in the optimized embeddings}
   \label{alg:Encoding_multi-scale}
\begin{algorithmic}
   \STATE {\bfseries Input:} Set of points $\left \{ \mathbf{x}_{i} \right \}_{i=1}^{N}$.
    \STATE  {\bfseries Output:} 
    Initial Node Embeddings  ${\psi}_{ \mathbf{x}}  $ . 
   \STATE \textbf{Step 1:} Construct $G=(V,\mathbf{W})$ from $\left \{ \mathbf{x}_{i} \right \}_{i=1}^{N}$.  
   \STATE \textbf{Step 2:} Construct the Laplacian $\mathbf{L}$. 
   \STATE \textbf{Step 3:} Compute $\lambda_{\mbox{max}}(\mathbf{L})$. 
   \STATE \textbf{Step 4}: \textbf{Initial Embedding construction:}  \\
  For $r = 1,..,D $ associated with the feature signals $  \left \{ f_{r} \right \}$ do: \\
Compute $\psi_{ {f}_{r}}(s_j,:)$ at scales $s_{j}, j=1,...K$ using Chebyshev approximation. \\
\STATE \textbf{Step 5}: Concatenate $\psi_{ {f}_{r}}(s_j,:)$ for  $r=1,...,D$ in a matrix form  $\psi_{ \mathbf{x}}(s_{j},:,:)$ for each fixed scale $s_{j}$, representing all scales using 3D tensor $ \psi_{\mathbf{x}} \in \mathbb{R}^{K} \times\mathbb{R}^{D}\times\mathbb{R}^{N}$. 
\end{algorithmic}
\end{algorithm} 

\subsection{Step 2: Optimization design using multi-scale network structure.}

We propose to use optimization based on SGD that begins with the initial embedding within the spectral graph domain. While employing positive and negative sampling, our aim is to optimize the embedding space directly in the SGW domain, revealing the intrinsic structure of manifold data while retaining high-frequency information associated with the graph Laplacian. This optimization within the SGW domain provides more effective regularization, allowing for the direct removal of noise from each spectral band. We outline the steps of the proposed approach.\\
\textbf{3D-Tensor Based Optimization for graph Embedding} Encoding the representation in a 3D tensor offers two key advantages. First, is the computational efficiency: the update rule at each iteration, associated with a pair of nodes, is performed jointly and simultaneously for all scales. This avoids the need for independent optimization per scale, which would significantly increase computational complexity. The second is optimization alignment: jointly optimizing across scales ensures alignment in the optimization process for all scales with respect to a fixed pair of nodes (positive or negative sample). This approach is more effective than optimizing scales independently, as it maintains consistency and improves overall performance. While SGW captures global graph structure, each embedding update remains feature-specific, preserving the one-to-one correspondence between input features and embedding dimensions, which is essential for interpretability. The encoded manifold representation is a 3D tensor $\psi_{\mathbf{x}} \in \mathbb{R}^{K} \times \mathbb{R}^{D} \times \mathbb{R}^{N}$. \\
An SGD update rule at iteration $t$, applied to the 3D tensor, is:

\begin{equation}
\label{Eq:SGD_update}
(\tilde{\psi}_{ \mathbf{x}}^{(t+1)})_{i,j,k} = 
(\tilde{\psi}_{ \mathbf{x}}^{(t)})_{i,j,k} - \alpha \frac{\partial \mathcal{L}} {\partial (\psi_{ \mathbf{x}})_{i,j,k}}
\end{equation}

where $\alpha$ is the learning parameter. In this case, we employ the cross-entropy loss function:

\begin{align}
\label{Eq:cross_en}
    \mathcal{L}(\tilde{\psi}_{ \mathbf{x}}|\mathbf{W}) 
    =  \sum_{i,j,k} &\left( w_{ij} \log \frac{w_{ij}}{v_{ij}^{\psi_{ \mathbf{x}} (s_{k}, :, :) }} \right. \nonumber \\
    &\left. + (1 - w_{ij}) \log \frac{1 - w_{ij}}{1 - v_{ij}^{\psi_{ \mathbf{x}} (s_{k}, :, :)}} \right)
\end{align}
where $ v_{ij}^{\psi_{ \mathbf{x}}(s_k,:,:
)}  = \frac{1}{1 + ||\psi_{ \mathbf{x}_i}(s_k,:,:
) - \psi_{ \mathbf{x}_j}(s_k,:,:
) ||^{2} }$. \\
We apply the optimization based on SGD with respect to each scale $s_{k}$, where the gradient of the loss is approximated by:

\begin{align}
\label{Eq:loss_derv}
   \frac{\partial \mathcal{L} \left(\tilde{\psi}_{\mathbf{x}}(s_{k}, :, :) \mid \mathbf{W}\right)} 
   {\partial \psi_{\mathbf{x}_i}(s_{k}, :, :)} 
   &= \sum_{j} w_{ij} v_{ij}^{\psi_{\mathbf{x}}(s_{k}, :, :)} \, r_{ij}(s_{k}) \\
   &\quad - \sum_{j} \frac{1}{\| r_{ij}(s_{k}) \|^{2}} v_{ij}^{\psi_{\mathbf{x}}(s_{k}, :, :)} \, r_{ij}(s_{k})
\end{align}
where $r_{ij}(s_{k}) = \psi_{ \mathbf{x}_i}(s_{k},:,:) - 
   \psi_{ \mathbf{x}_j}(s_{k},:,:)$.  \\
   \textbf{Feature updates:} Our implementation employs an optimization process that operates jointly across all scales using the 3D tensor, as described in the SGD update rule for iteration $t$ in Eq.(\ref{Eq:SGD_update}). The optimization process uses SGD with positive and negative sampling for all scales $s_{k}$ given a pair of positive or negative pair of nodes $i,j$. At each iteration $t$, associated with either a positive sample (an edge) or a negative sample (a pair of unconnected nodes), the update rule from Eq.(\ref{Eq:SGD_update}) is applied. The derivative of the loss function, as detailed in Eq.(\ref{Eq:loss_derv}), is computed for all scales $(s_1,... s_K)$, with the sign in Eq.(\ref{Eq:SGD_update}) depending on whether the sample is positive or negative. Once the optimization concludes, the final embedding is derived from the 3D tensor by summing over the scales, as specified in Eq.(\ref{Eq:sum_wavelets_scales}). Recall that the term $\psi_{\mathbf{x}}(s_k, :, :) \in \mathbb{R}^{D \times N}$ is a slice of the 3D SGW tensor at scale $s_k$, representing spectral graph wavelet coefficients for all features across all nodes $N$. The specific vector $\psi_{\mathbf{x}}(s_k, j, :) \in \mathbb{R}^N$ corresponds to the SGW coefficients at scale $s_k$ for input feature $j$ across all nodes. Note that in Eq.(\ref{Eq:SGD_update}), the $j$-th embedding coordinate—aligned with input feature $j$—is updated via SGD using two components: (1) the current embedding value and (2) a structural term involving $v_{ij}^{\psi_{\mathbf{x}}(s_k, :, :)}$, which integrates the graph structure and multiscale information from the SGW tensor. At each iteration, updates for positive/negative node pairs are applied jointly across all scales, but each scale's contribution is weighted by its energy concentration—ensuring that more informative scales have greater influence. We compute the final embedding by summing up all the optimized embeddings with respect to each scale $s_{k}$
\begin{equation}
\label{Eq:sum_wavelets_scales}
    \tilde{\psi}_{\mathbf{x}} = \sum_{k}  \tilde{\psi}_{\mathbf{x}}(s_{k}, :,:)
\end{equation} 
The dimensionality of the final embedding is $\tilde{\psi}_{\mathbf{x}}\in \mathbb{R}^{D}\times  \mathbb{R}^{N}$, maintaining the same dimensionality as the original input, and thus there exists a one-to-one correspondence between the coordinates of the input features \(\mathbf{x}\) and \(\tilde{\psi}_{\mathbf{x}}\). This correspondence ensures that each dimension of the feature space is directly mapped to a single dimension in the embedding space \(\tilde{\psi}_{\mathbf{x}}\). Our optimization process involves using SGD with positive and negative sampling, similar to recent graph embeddings methods (\cite{Largevis, UMAP}). Positive edge samples are associated with attraction (first term on the right hand side), while negative samples (second term on the right hand side) refer to a pair of nodes that are not connected in the graph, which create repulsion among dissimilar points. \\
\textbf{Remark 1:} While we adopt the SGW formulation by Hammond et al. (\cite{Hammond}), our framework is compatible with other spectral graph wavelet and multiscale representations. \\
\textbf{Remark 2:}  While the manifold hypothesis suggests the data lie near a lower-dimensional manifold, this does not preclude learning meaningful structure in the ambient space. Our method incorporates high-frequency graph information via spectral graph wavelets, enhancing stability and local structure capture, without requiring expensive eigenspectrum computations as in Laplacian Eigenmaps or UMAP. 
 \section{Feature Importance using Explicit Correspondence with Embedding Features}
\label{sec:Interpretable Embedding}
Feature importance is one of the simplest yet most fundamental tools for model interpretability. It associates input features with model outputs, highlighting which features most strongly influence the learned representations. We show that feature subsets selected from the MS-IMAP embedding lead to superior clustering performance compared to those selected from the input space using the same feature importance methods. Unlike competing approaches, MS-IMAP retains semantic alignment and supports direct correspondence between embedding dimensions and original input features. This makes MS-IMAP an interpretable alternative to black-box models. We empirically demonstrate that applying feature importance techniques to the MS-IMAP embedding yields more informative feature subsets for clustering than applying the same techniques directly to the raw input. This provides evidence of the semantic consistency preserved by our method—often absent in traditional nonlinear embeddings. We explore two alternative approaches for estimating feature importance, leveraging the explicit correspondence in the embedding space where each dimension maps back to a specific input feature. These methods are evaluated in an unsupervised or self-supervised setting, where feature selection is performed without access to label information.\\
\subsection{Assessing Embedding Feature Importance using the Laplacian Score:} The first method utilizes the Laplacian Score (LS) (\cite{Ls}), which assesses the importance of each feature by examining its correlation with the eigenvalues of the graph Laplacian. The significance of the variable corresponding to the \( l \)-th coordinate, represented by the feature \( f_l \in \mathbb{R}^N \), is inferred from the relevance of its corresponding coordinate \( l \) in the embedding space, \( \tilde{\psi}_{\mathbf{x}_l} \).  
\textbf{Computing the Laplacian score with respect to the embedding features  \( \tilde{\psi}_{\mathbf{x}} \)}. 
The Laplacian score is calculated with respect to embedding features \( \tilde{\psi}_{\mathbf{x}} \) using the Laplacian graph \( \mathbf{L} \) and the degree matrix \( \mathbf{D} \). To compute the Laplacian Score, we first subtract the mean and then calculate it as follows:
$$ L_{s}(\tilde{\psi}_{\mathbf{x}})_l = \frac{(\tilde{\psi}_{\mathbf{x}})_l^{T} \mathbf{L}(\tilde{\psi}_{\mathbf{x}})_l}{(\tilde{\psi}_{\mathbf{x}})_l^{T} \mathbf{D}(\tilde{\psi}_{\mathbf{x}})_l}$$
Smaller scores indicate that the embedding feature $(\tilde{\psi}_{\mathbf{x}})_l$ has a greater projection onto the subspace of eigenvectors corresponding to the smallest eigenvalues, signifying higher importance concerning the global graph structure. Therefore, the feature importance of $(\tilde{\psi}_{\mathbf{x}})_l$ can be directly interpreted as the importance of the corresponding original feature. Although the graph is typically constructed using local neighborhoods (e.g., via k-nearest neighbors), the Laplacian operator reflects global smoothness over the entire data manifold. 

\subsection{Assessing Importance of Embedding Features using Mutual Information} We also propose a self-supervised approach that uses Mutual Information (MI) to estimate the importance of each dimension in the embedding space, which can then be explicitly linked to the original features. First, we apply $k$-means to cluster the MS-IMAP embedding space and use the resulting clusters as pseudo-labels. We then compute MI to quantify the shared information between features and these pseudo-labels, similar to how MI would be computed with ground truth labels. This step allows MI to measure how much each feature contributes to cluster assignments, highlighting those that are the most informative to distinguish clusters. The MI-Based Feature Importance identify features that are informative about cluster assignments. Thus, masking these features would especially degrade the clustering performance. We summarize the MI-based unsupervised feature selection in Algorithm \ref{alg:MI-based}.

\subsection{Comparison: Laplacian Score (LS) vs. Feature Importance Based-Mutual Information}

\begin{algorithm}[tb]
   \caption{MI-based Unsupervised Feature Selection}
   \label{alg:MI-based}
\begin{algorithmic}
   \STATE {\bfseries Input:} MS-IMAP embeddings  $ \tilde{\psi}_{\mathbf{x}}$, number of clusters $k$
   \STATE {\bfseries Output:} MI scores for each feature
   
   \STATE \textbf{Step 1:} Apply k-means clustering on the embedding space $\psi_{\mathbf{x}}$ to obtain cluster labels
   \STATE \hspace{1cm} $\text{clusters} = \text{k-means}(  \tilde{\psi}_{\mathbf{x}}, k)$
   
   \STATE \textbf{Step 2:} Use the resulting cluster labels as pseudo-labels
   \STATE \hspace{1cm} $\text{pseudo\_labels} = c(  \tilde{\psi}_{\mathbf{x}})$
   \STATE \textbf{Step 3:} Initialize an empty list to store mutual information (MI) values for each feature
   \STATE \hspace{1cm} $\text{mi\_scores} = []$
   
   \STATE \textbf{Step 4:} For each feature $i$ in the original data or embedding space:
   \STATE \hspace{1cm} \textbf{Step 4.1:} Compute mutual information between feature $  \tilde{\psi}_{\mathbf{x}}(i)$ and the pseudo-labels
   \STATE \hspace{1cm} \textbf{Step 4.2:} $MI\_score = \text{MI}(  \tilde{\psi}_{\mathbf{x}}(i), c(  \tilde{\psi}_{\mathbf{x}})$)
   
   \STATE \hspace{1cm} \textbf{Step 4.3:} Append $MI\_score$ to $\text{mi\_scores}$
   
   \STATE \textbf{Step 5:} Output the list of MI scores
   \STATE \hspace{1cm} \textbf{Return} $\text{mi\_scores}$
\end{algorithmic}
\end{algorithm}

\begin{figure}[t]
  \centering
  \begin{subfigure}{0.4\columnwidth}
    \centering
    \includegraphics[width=\linewidth]{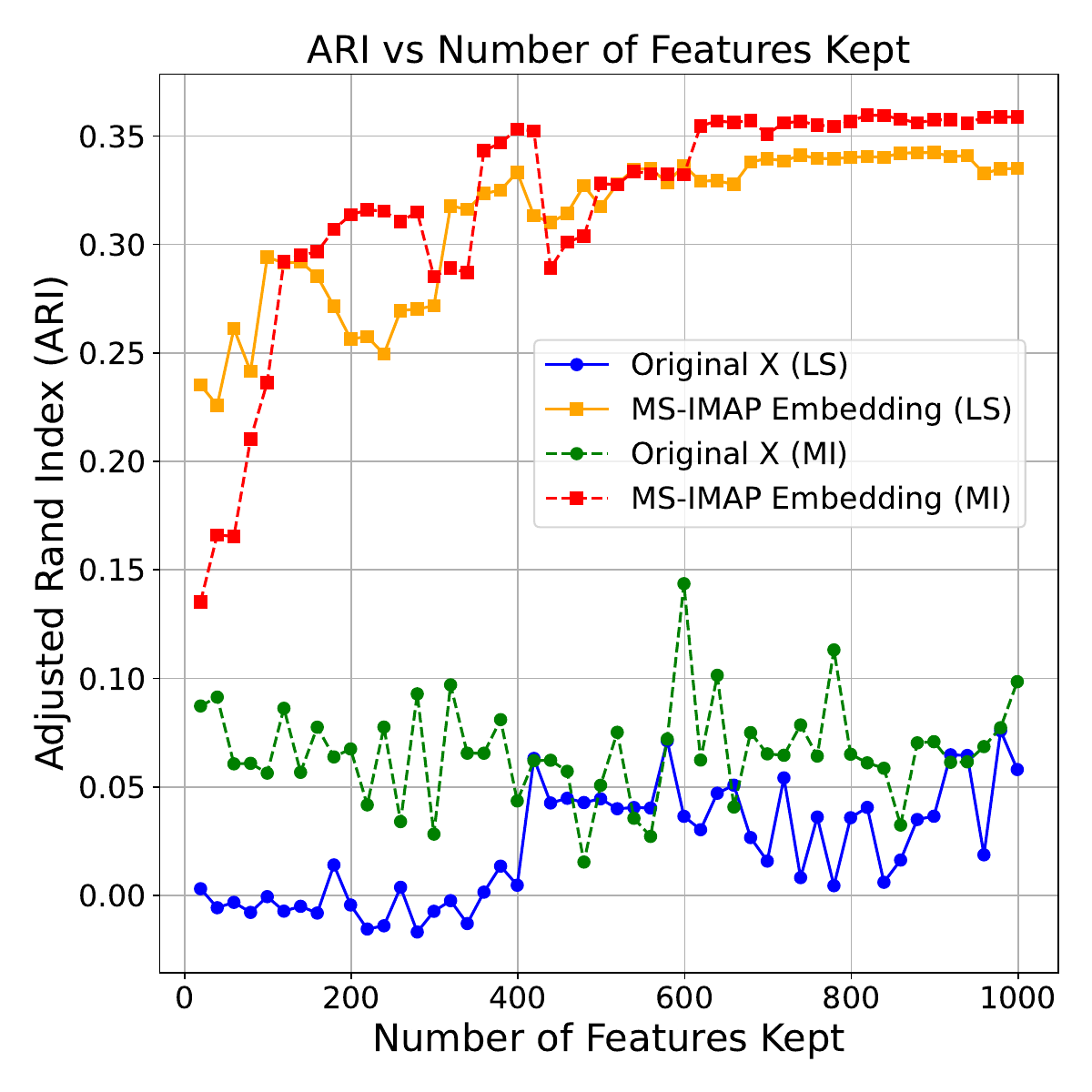}
    \caption{Cora dataset (ARI)}
    \label{fig:feat_import_cora_ARI}
  \end{subfigure}
  \hspace{0.05\columnwidth}
  \begin{subfigure}{0.4\columnwidth}
    \centering
    \includegraphics[width=\linewidth]{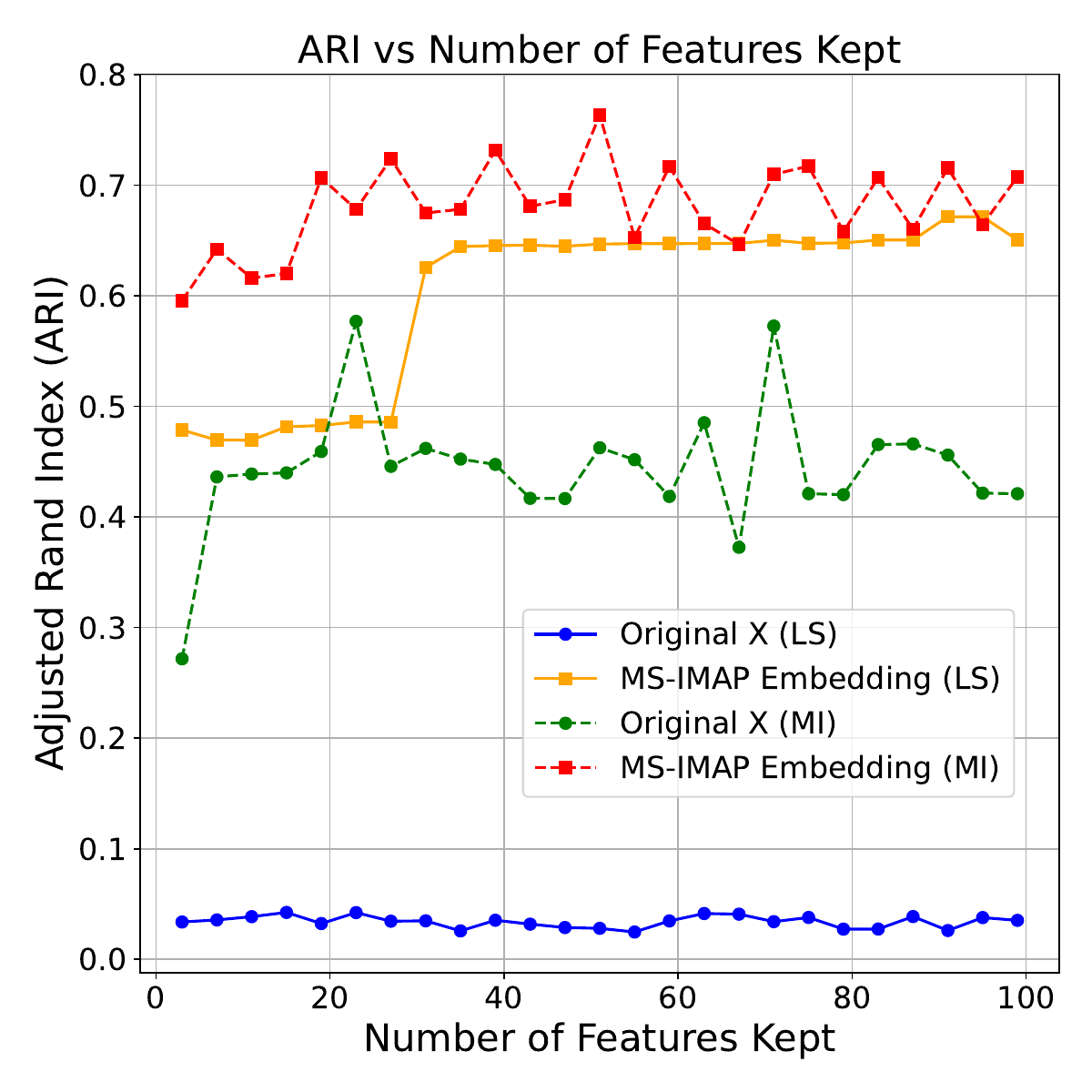}
    \caption{Zillions dataset (ARI)}
    \label{fig:feat_import_zillions_ARI}
  \end{subfigure}
  \caption{Clustering results using Adjusted Rand Index (ARI) on Cora and Zillions datasets with a subset of features based on feature importance, comparing the MS-IMAP embedding space and the input features.}
\end{figure}

\begin{figure}[t]
  \centering
  \begin{subfigure}[b]{0.4\columnwidth}
    \centering
    \includegraphics[width=\linewidth]{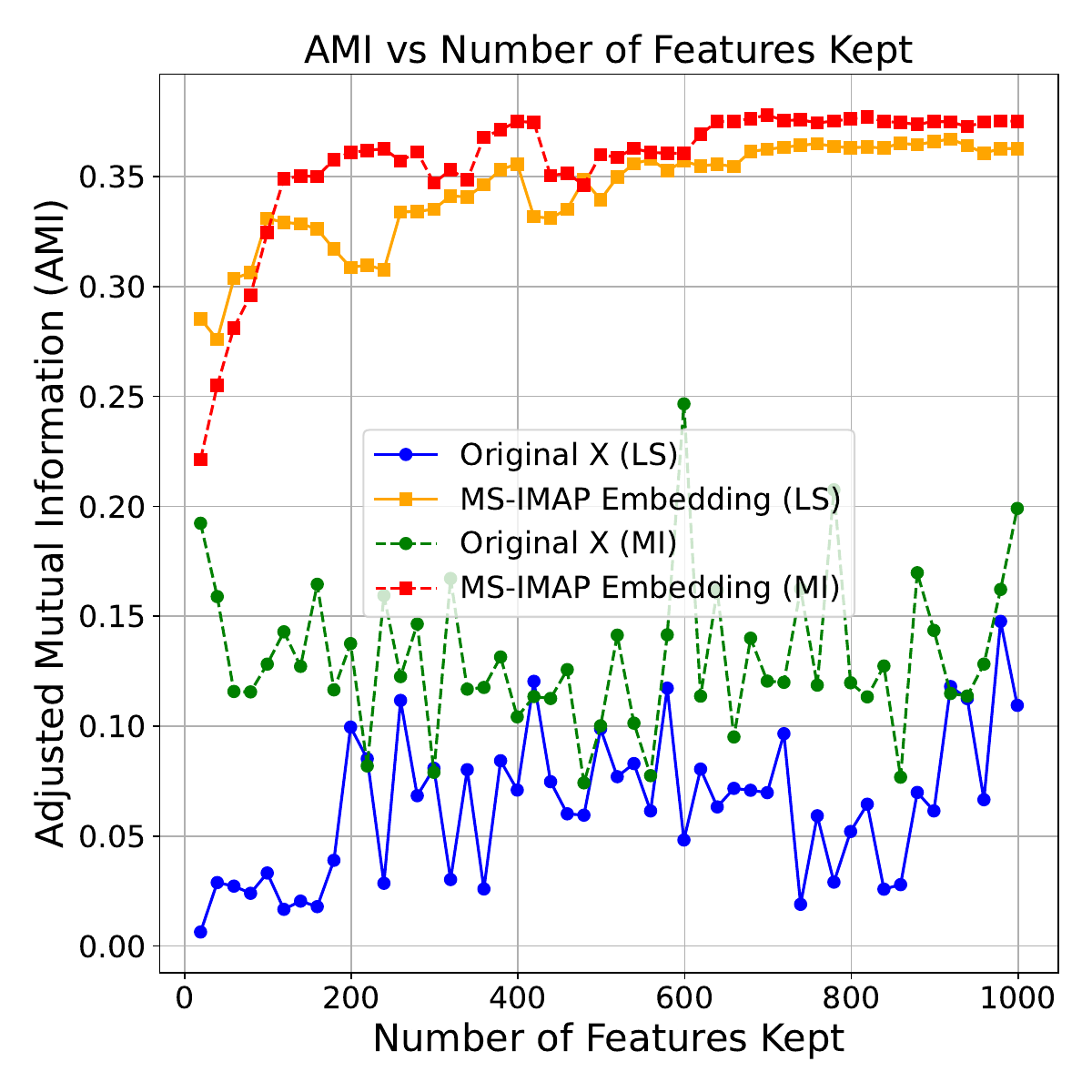}
    \caption{Cora dataset (AMI)}
    \label{fig:feat_import_cora_AMI}
  \end{subfigure}
  \hspace{0.05\columnwidth}
  \begin{subfigure}[b]{0.4\columnwidth}
    \centering
    \includegraphics[width=\linewidth]{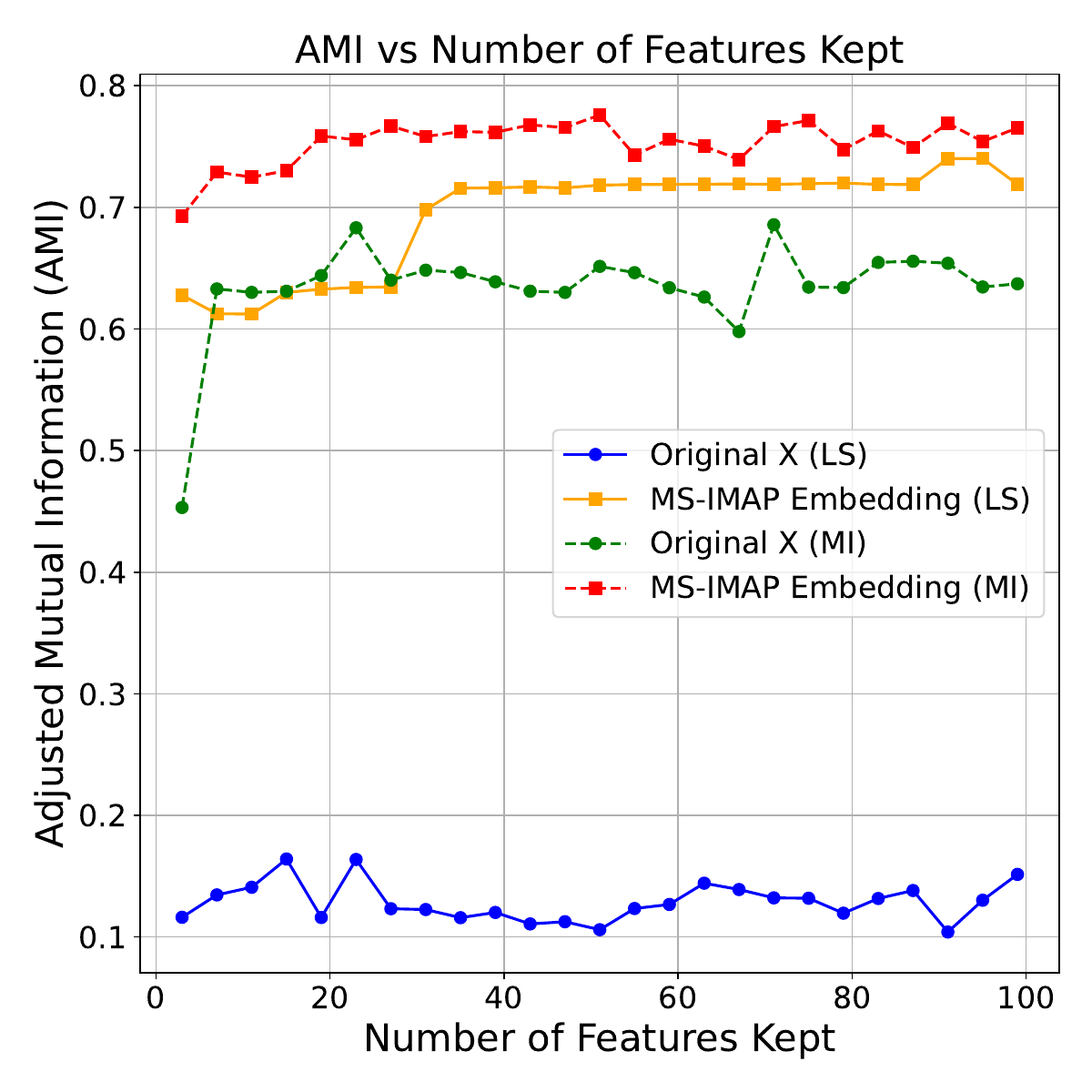}
    \caption{Zillions dataset (AMI)}
    \label{fig:feat_import_zillions_AMI}
  \end{subfigure}
  \caption{Clustering results using Adjusted Mutual Information (AMI) on Cora and Zillions datasets with a subset of features based on feature importance, comparing the MS-IMAP embedding space and the input features.}
  \label{fig:feat_import_cora_zillions_all}
\end{figure}

The Laplacian Score (LS) and Mutual Information (MI)–based feature importance are two complementary approaches to unsupervised feature selection, with LS focusing on geometric smoothness over a graph and MI targeting feature relevance with respect to clustering structure.
\textbf{Laplacian Score (LS)} evaluates how smoothly a feature varies over a graph constructed from the data. It is a \textbf{fully unsupervised}, geometry-driven criterion that does not require clustering or pseudo-labels. A feature receives a low score if it varies slowly and consistently across the graph. This makes LS particularly effective for identifying features that align with smooth manifold structure. However, LS also has several limitations. It is not task-aware: it does not assess whether a feature is relevant for a specific downstream objective such as clustering or classification. Moreover, LS can overrate irrelevant but structured noise—for instance, a smoothly varying but non-informative feature may receive a deceptively low score simply because it is smooth. \\
In contrast, \textbf{Mutual Information–based feature importance} measures how much information a feature provides about cluster assignments. It quantifies the statistical dependency between each feature and the resulting clusters (e.g., from k-means or MS-IMAP embeddings). A key strength of MI is that it is task-aware: it is self-supervised (or can be seen as supervised with respect to pseudo-labels), capturing which features are actually useful for distinguishing between clusters. MI scores are also directly interpretable—a high MI score indicates that the feature contributes to separating groups—and the method is generally robust to irrelevant noise features, which receive low MI if they do not correlate with clustering structure. However, MI is sensitive to the quality of the clustering used to generate pseudo-labels—poor or unstable clustering can diminish its effectiveness.

\section{Experimental Results}
\label{sec:Experimental_Results}

We evaluated our approach in both synthetic and real datasets in two application domains. First, we demonstrate the use of MS-IMAP embeddings for feature importance—a novel application that highlights the interpretability of our method. Next, we assess the performance of MS-IMAP in clustering tasks using the full embedding space, comparing it against a range of manifold learning-based embedding techniques. \\
\textbf{Datasets} We study the performance of MS-IMAP compared to other methods for real datasets. We chose a mix of datasets from varied fields: the Census dataset (\cite{UCI}) is a financial dataset containing information about individuals extracted from the 1994 US Census; the Zilionis dataset (\cite{Zilionis}) is a biological dataset containing single-cell sequencing data from different types of cells; the Animals with Attributes (AWA) (\cite{Xian2017ZeroShotL}) dataset is an image dataset containing images of animals; The Human Activity Recognition (HAR) (\cite{HAR}) dataset which consists of sensor data collected from subjects performing six different activities:; and the Cora dataset (\cite{sen2008collective, mccallum2000automating}) is a graph dataset of a citation network consisting of scientific publications. Note that we only use the features without the natural graph associated with Cora (see more information about each dataset in the Appendix). 

\subsection{Assessing Feature Importance via Embedding Correspondence}
\label{sec:Assessing Feature Importance}
We demonstrate the application of MS-IMAP for feature importance selection by evaluating clustering performance on subsets of the most significant features. Feature selection is performed with respect to both the input space and the MS-IMAP embedding space, using two alternative unsupervised methods: Laplacian Score and Mutual Information-based selection. We compare clustering results obtained by applying \$k\$-means to subsets of features selected from the embedding, against those selected directly from the original feature space using the same methods. Unlike most manifold learning and graph embedding techniques, MS-IMAP produces embeddings in which each dimension corresponds directly to an input feature. This one-to-one mapping enables direct assessment of individual feature contributions, preserving interpretability. \\
\textbf{Evaluation metrics:} Clustering performance is measured using Adjusted Rand Index (ARI) and Adjusted Mutual Information (AMI). In all experiments, $k$-means is applied in the embedding space. On the Cora and Zillions datasets, Figures \ref{fig:feat_import_cora_ARI}, \ref{fig:feat_import_zillions_ARI}, \ref{fig:feat_import_cora_AMI}, and \ref{fig:feat_import_zillions_AMI} show that, for both feature selection methods and for both evaluation metrics, clustering accuracy is consistently higher when using the MS-IMAP embedding compared to the original features—across varying feature subset sizes. This improvement highlights the robustness of the MS-IMAP embedding, while maintaining interpretability through its feature-wise alignment. Additional experiments on the AWA and HAR datasets, presented in the Appendix, further confirm these gains.

\subsection{Clustering Output Embedding Space}
We evaluate our method by testing its clustering performance using the full output embedding and comparing it to several representative methods, including UMAP, t-SNE, Diffusion Maps, ISOMAP, PHATE (\cite{moon2019visualizing}), PaCMAP (\cite{Pacmap}), TriMap (\cite{trimap}) and HeatGeo (\cite{NEURIPS2023_16063a1c}). While most of these methods perform dimensionality reduction, we emphasize that dimensionality reduction is a methodological design choice, not an inherent advantage. Many manifold learning techniques default to low-dimensional embeddings (e.g., 2D or 3D) due to computational constraints such as eigendecomposition, yet this often comes at the cost of reduced expressiveness and, particularly for nonlinear methods, diminished interpretability. However, since most baseline manifold learning methods we compare rely on nonlinear dimensionality reduction, we include clustering results based on the embedding space of the Graph Scattering Transform (\cite{Scattering_Graph}) for completeness. The Graph Scattering Transform, which is primarily used in graph classification and semi-supervised learning tasks, can be adapted for unsupervised learning via simple vector-based concatenation. We note that the Graph Scattering Transform offers less interpretability than MS-IMAP, which preserves a one-to-one correspondence between input features and embedding dimensions. Additionally, the increased dimensionality of the embeddings produced by the Graph Scattering Transform can lead to higher computational costs. As shown in Table \ref{tab:ARI_AMI}, MS-IMAP consistently delivers stable and competitive performance, ranking best or second-best across all benchmarks. In contrast, state-of-the-art methods often underperform or show instability on at least one dataset. In addition, MS-IMAP provides an interpretable embedding through its one-to-one correspondence with the original features—unlike the other competing graph embedding methods. \\
\textbf{Runtime and Computational Complexity} The computational complexity of Multi-Scale IMAP is of \\ $O(ND log(N))$. Experiments were performed on Virtual Server containers with 32 cores Intel Xeon 8259CL running at 2.50Ghz and 256GB of RAM. Our method is practical for handling datasets with millions of points and dozens to hundreds of features within a few hours, making the method well-suited for real-world applications.  For example, on the Cancer QC data-set of 48,969 samples and 306 features took 27 mins. Additional details and running times on all datasets are provided in the Appendix.  Note that the increase in running times compared to methods such as UMAP is primarily due to our extension of manifold learning to include interpretability. 

\begin{table*}[htbp]
\centering
\resizebox{\textwidth}{!}{%
\begin{tabular}{lllllllllll}
\toprule
Data              & \multicolumn{2}{c}{HAR} & \multicolumn{2}{c}{Census} & \multicolumn{2}{c}{Zilionis} & \multicolumn{2}{c}{AWA} & \multicolumn{2}{c}{Cora} \\
\midrule
Method       & ARI & AMI & ARI & AMI & ARI & AMI & ARI & AMI & ARI & AMI \\
\midrule
Features     & 0.45$\pm$0.00 & 0.58$\pm$0.00 & 0.11$\pm$0.00 & 0.15$\pm$0.00 & 0.48$\pm$0.04 & 0.63$\pm$0.01 & 0.07$\pm$0.00 & 0.16$\pm$0.00 & 0.01$\pm$0.02 & 0.07$\pm$0.02 \\
ISOMAP       & 0.55$\pm$0.00 & 0.65$\pm$0.00 & 0.18$\pm$0.00 & 0.09$\pm$0.00 & 0.44$\pm$0.03 & 0.56$\pm$0.00 & 0.52$\pm$0.00 & 0.60$\pm$0.00 & 0.12$\pm$0.02 & 0.17$\pm$0.02 \\
DM           & 0.30$\pm$0.00 & 0.49$\pm$0.00 & 0.01$\pm$0.00 & 0.02$\pm$0.00 & \multicolumn{1}{c}{-} & \multicolumn{1}{c}{-} & 0.22$\pm$0.00 & 0.42$\pm$0.00 & 0.06$\pm$0.00 & 0.07$\pm$0.00 \\
UMAP         & 0.61$\pm$0.00 & 0.73$\pm$0.00 & 0.23$\pm$0.00 & 0.15$\pm$0.00 & 0.52$\pm$0.00 & 0.71$\pm$0.00 & \textbf{0.74}$\pm$0.00 & \textbf{0.82}$\pm$0.00 & 0.33$\pm$0.01 & \textbf{0.36}$\pm$0.02 \\
t-SNE        & 0.62$\pm$0.00 & 0.70$\pm$0.00 & 0.20$\pm$0.00 & \textbf{0.24}$\pm$0.00 & 0.39$\pm$0.01 & 0.69$\pm$0.00 & 0.71$\pm$0.00 & 0.80$\pm$0.00 & 0.20$\pm$0.00 & 0.24$\pm$0.00 \\
Scattering.T & 0.45$\pm$0.00 & 0.58$\pm$0.00 & 0.12$\pm$0.00 & 0.16$\pm$0.00 & 0.46$\pm$0.04 & 0.64$\pm$0.02 & 0.67$\pm$0.00 & 0.75$\pm$0.00 & 0.06$\pm$0.03 & 0.14$\pm$0.03 \\
PHATE        & 0.55$\pm$0.02 & 0.66$\pm$0.02 & 0.20$\pm$0.00 & 0.09$\pm$0.01 & 0.65$\pm$0.01 & 0.72$\pm$0.00 & 0.66$\pm$0.03 & 0.73$\pm$0.00 & 0.12$\pm$0.01 & 0.22$\pm$0.01 \\
TriMap       & 0.59$\pm$0.02 & 0.72$\pm$0.01 & 0.20$\pm$0.00 & 0.14$\pm$0.00 & 0.62$\pm$0.08 & \textbf{0.77}$\pm$0.01 & 0.74$\pm$0.04 & 0.81$\pm$0.01 & 0.15$\pm$0.00 & 0.20$\pm$0.00 \\
PaCMAP       & 0.62$\pm$0.01 & 0.75$\pm$0.08 & \textbf{0.24}$\pm$0.00 & 0.15$\pm$0.00 & 0.55$\pm$0.03 & 0.76$\pm$0.00 & 0.75$\pm$0.04 & 0.81$\pm$0.01 & 0.21$\pm$0.01 & 0.26$\pm$0.00 \\
HeatGeo      & 0.60$\pm$0.00 & 0.69$\pm$0.00 & 0.15$\pm$0.00 & 0.10$\pm$0.00 & \multicolumn{1}{c}{-} & \multicolumn{1}{c}{-} & 0.65$\pm$0.00 & 0.74$\pm$0.00 & 0.17$\pm$0.00 & 0.20$\pm$0.00 \\
MS-IMAP      & \textbf{0.67$\pm$0.01} & \textbf{0.80$\pm$0.01} & 0.23$\pm$0.00 & 0.15$\pm$0.00 & \textbf{0.70$\pm$0.01} & 0.76$\pm$0.01 & \textbf{0.74$\pm$0.00} & 0.81$\pm$0.00 & \textbf{0.33$\pm$0.00} & \textbf{0.36$\pm$0.00} \\
\bottomrule
\end{tabular}
}
\caption{Clustering results comparison using ARI and AMI on the Census, Zilionis, AWA, and Cora datasets. Each entry report the mean clustering accuracy and standard deviation. The best performance is \textbf{bolded}.}
\label{tab:ARI_AMI}
\end{table*}

\section{Theoretical results: Sampling set for Smooth Manifolds with functions defined over Paley-Wiener Spaces} 

In this section, we characterize the theoretical properties of the representation power of the SGW operator by considering functions sampled from the Paley-Wiener spaces \cite{Pesenson2008SamplingIP} on combinatorial graphs. Pesenson introduced the Paley-Wiener spaces and demonstrated that Paley-Wiener functions of low type are uniquely determined by their values on certain subsets, known as uniqueness sets $U$. We show that the SGW operator can represent functions \( f \) within the Paley-Wiener space more efficiently than the graph Laplacian operator $\mathbf{L}$. This efficiency is demonstrated by showing that the SGW operator is more effective in representing functions with larger bandwidth \( \omega \) in the Paley-Wiener spaces (i.e., with higher frequencies) using subsets of nodes from the uniqueness sets \( U \). To characterize the representation properties of functions defined over \( PW_{\omega}(G) \), we employ the notion of the \( \Lambda \)-set, introduced by Pesenson which is central to our investigation. Formally, the Paley-Wiener space of \( \omega \)-bandlimited signals is defined as \( PW_{\omega}(G) = \left\{ f \mid \hat{f}(\lambda) = 0 \,\, \forall \,\, \lambda > \omega \right\} \). The space \( L_{2}(G) \) is defined as the Hilbert space of all complex-valued functions, and \( L_{2}(S) \) is defined as the space of all functions from \( L_{2}(G) \) with support in \( S \): \( L_{2}(S) = \left\{ \varphi \in L_{2}(G) \mid \varphi(v) = 0, v \in V(G) \setminus S \right\} \). The \( \Lambda \)-set is defined as follows: a set of vertices \( S \subset V \) is a \( \Lambda(S) \)-set if any \( \varphi \in L_{2}(S) \) satisfies the inequality  $ ||\varphi|| \leq \Lambda ||\mathbf{L} \varphi|| $ for some constant \( \Lambda(S) > 0 \). The infimum of all \( \Lambda > 0 \) for which \( S \) is a \( \Lambda \)-set is denoted by \( \Lambda \). The ability of the SGW operator to efficiently represent functions \( f \in PW_{\omega}(G) \) can be summarized in the following theorem, which highlights the role of the \( \Lambda_{\psi} \)-set with respect to the operator \( \psi \). We show that any signal \( f \in PW_{\omega}(G) \), where \( \lambda_{1} \leq \omega < \Omega_{G} \) for some \( \Omega_{G} < \lambda_{N} \), can be uniquely represented by its samples on the uniqueness set \( U \) using the SGW operator. Under certain conditions related to the SGW operator, its associated \( \Lambda_{\psi} \)-set is smaller than the \( \Lambda \)-set associated with the Laplacian operator.\\
\textbf{Theorem 3}
Let \( G = (V, \mathbf{W}) \) be a connected graph with \( n \) vertices and \( f \in PW_{\omega}(G) \) for \( \lambda_{1} \leq \omega < \lambda_{\text{max}} \). The SGW operator \( \psi \) can be constructed such that the set \( S \) is a \( \Lambda_{\psi} \)-set and the set \( U = V \setminus S \) is a uniqueness set for any space \( PW_{\omega}(G) \) with \( \omega < 1/\Lambda_{\psi} \) and \( \Lambda_{\psi} < \Lambda \) for any \( \varphi \in L_{2}(S) \), where \( \Lambda \) is the \( \Lambda \)-set of the Laplacian operator. \\\\
The proof is provided in the Appendix (see Theorem 3 in \ref{sec:Theoreticalresultsproofs}). Theorem 3 motivates our use of SGW within the proposed framework. Specifically, it shows that SGW operators provide more compact and effective representations of band limited functions in Paley-Wiener spaces on graphs than the traditional Laplacian operator. 

\section{Discussion}
Identifying the key factors in high-dimensional datasets is essential for leveraging unlabeled data in many practical applications. In this work, we introduce a novel contrastive learning framework for interpretable manifold learning via graph embeddings, leveraging both low- and high-frequency information to enhance representation quality. Our method employs Spectral Graph Wavelets (SGW) to construct a multi-scale graph representation of the input feature space. This representation is optimized using a stochastic gradient descent (SGD)-based scheme, combined with a novel 3D tensor encoding strategy that enriches the embedding process.

MS-IMAP differs fundamentally from traditional dimensionality reduction or visualization techniques. While many manifold learning methods (e.g., Laplacian Eigenmaps, Diffusion Maps, UMAP) aim to reveal intrinsic structure through low-dimensional embeddings, MS-IMAP targets a broader objective: constructing structure-preserving, interpretable embeddings suitable for downstream tasks. By design, MS-IMAP preserves input dimensionality and enforces a one-to-one correspondence between input features and embedding dimensions.

To enhance the embedding with structural information, we introduce a 3D tensor that captures SGW coefficients across nodes, features, and frequency bands. This enables the model to encode both global and local graph structure. Importantly, although SGW captures information across the graph, each embedding coordinate is updated solely based on the SGW coefficients of its corresponding input feature. This preserves the interpretability and separability of features throughout training. We demonstrate the utility of these interpretable embeddings for both feature selection and traditional clustering tasks. Specifically, we assess clustering performance based on features selected from both the input space and the embedding space, using two unsupervised methods: Laplacian Score and Mutual Information-based selection. A key advantage of our framework is its inherent linkage between the original and embedded feature spaces—a property rarely offered by existing nonlinear manifold learning techniques. Furthermore, we show that the resulting embeddings are highly competitive with state-of-the-art graph embedding methods.

To further justify the use of SGW for representation, we analyze its theoretical properties by studying functions in Paley-Wiener spaces on combinatorial graphs. We show that the SGW operator enables more expressive representations, characterized through the concept of the $\Lambda$-set. \\
\textbf{Limitations and future directions include several aspects.} First, our approach assumes that the input features—and the similarity measures used—adequately capture the underlying manifold structure, particularly geodesic distances. Second, extending the current SGD-based optimization framework to handle out-of-sample generalization remains a challenge. Finally, while the embedding-feature correspondence facilitates global feature importance analysis, further improvements may be achieved by incorporating localized importance estimation techniques.

\bibliography{main}
\bibliographystyle{plain}

\section{Appendix}
\subsection{Encoding Method 1}

In addition to the 3D-Tensor based encoding presented in the main body of the paper, one may also examine a simple encoding method that involves concatenating the multi-scale representation of all features and filters (associated with different scales) into a single vector representation for each point. This results in a matrix representation denoted as $\psi_{\mathbf{x}}$. We detail the encoding and optimization employed for this method (coined method 1). 
Note that we designate the concatenation using $\left |  \right |$, with

$$\mathbf{c} ( \psi_{{f}_{i}}(s_k,:),  \psi_{ {f}_{j}}(s_k,:) )  = \psi_{ {f}_{i}}(s_k,:) \left |  \right |  \psi_{ {f}_{j}}(s_k,:)$$  
denoting the concatenation of the vectors corresponding to the multi-scale representation $\psi_{{f}_{i}}(s_k,:)$ and $\psi_{{f}_{j}}(s_k,:)$. For method 1 all features and all scales are concatenated together, the resulting matrix $\psi_{\mathbf{x}}$ can be represented as
$$\psi_{\mathbf{x}} = \psi_{{f}_{1}}(s_1,:) \| \psi_{{f}_{2}}(s_1,:) \|.... \|\psi_{{f}_{D-1}}(s_K,:) \| \psi_{{f}_{D}}(s_K,:)$$ 
where $\psi_{\mathbf{x}} \in \mathbb{R}^{KD}\times\mathbb{R}^{N}$.

1.  \textbf{Optimize Embedding Method 1:}  \\
Given the encoded multi-scale representation $\psi_{\mathbf{x}}\in \mathbb{R}^{KD}\times\mathbb{R}^{N}$ perform optimization in the SGW domain, using the following fuzzy cross entropy loss function: 
\begin{align}
     \mathcal{L}(\tilde{\psi}_{ \mathbf{x}}|\mathbf{W})  = \sum_{i,j} 
 \left ( w_{ij} \mbox{log}\frac{w_{ij}}{v_{ij}^{\psi_{ \mathbf{x}}}} + (1- w_{ij}) \mbox{log}\frac{1 - w_{ij}}{1 - v_{ij}^{\psi_{ \mathbf{x}}}} \right )   
\end{align}

where $ v_{ij}^{\psi_{ \mathbf{x}}}  = \frac{1}{1 + ||\psi_{ \mathbf{x}_i} - \psi_{ \mathbf{x}_j} ||^{2} }$. 

Dropping terms that do not depends on $\psi_{ \mathbf{x}_{i}}$, the gradient of the loss is approximated by: 
\begin{align}
   \frac{\partial  \mathcal{L} (\tilde{\psi}_{ \mathbf{x}}|\mathbf{W})   } {\partial \psi_{ \mathbf{x}_i}}    \sim \sum_{j} w_{ij} v_{ij}^{\psi_{ \mathbf{x}}} ( r_{ij}) -  \sum_{j} 
 \frac{1}{||r_{ij} ||^{2} } v_{ij}^{\psi_{ \mathbf{x}}} (\psi_{ \mathbf{x}_i}- \psi_{ \mathbf{x}_j})
\end{align}
where $r_{ij} = \psi_{ \mathbf{x}_i} - 
   \psi_{ \mathbf{x}_j}$, and $\psi_{ \mathbf{x}_i}$ corresponds to the vector associated with node $i$ multi-scale representation concatenated for all the graph signals and all spectral bands.  since for method 1 all features and all scales are concatenated together, 
   
\begin{figure}
\centering
\includegraphics[width=0.7\columnwidth]{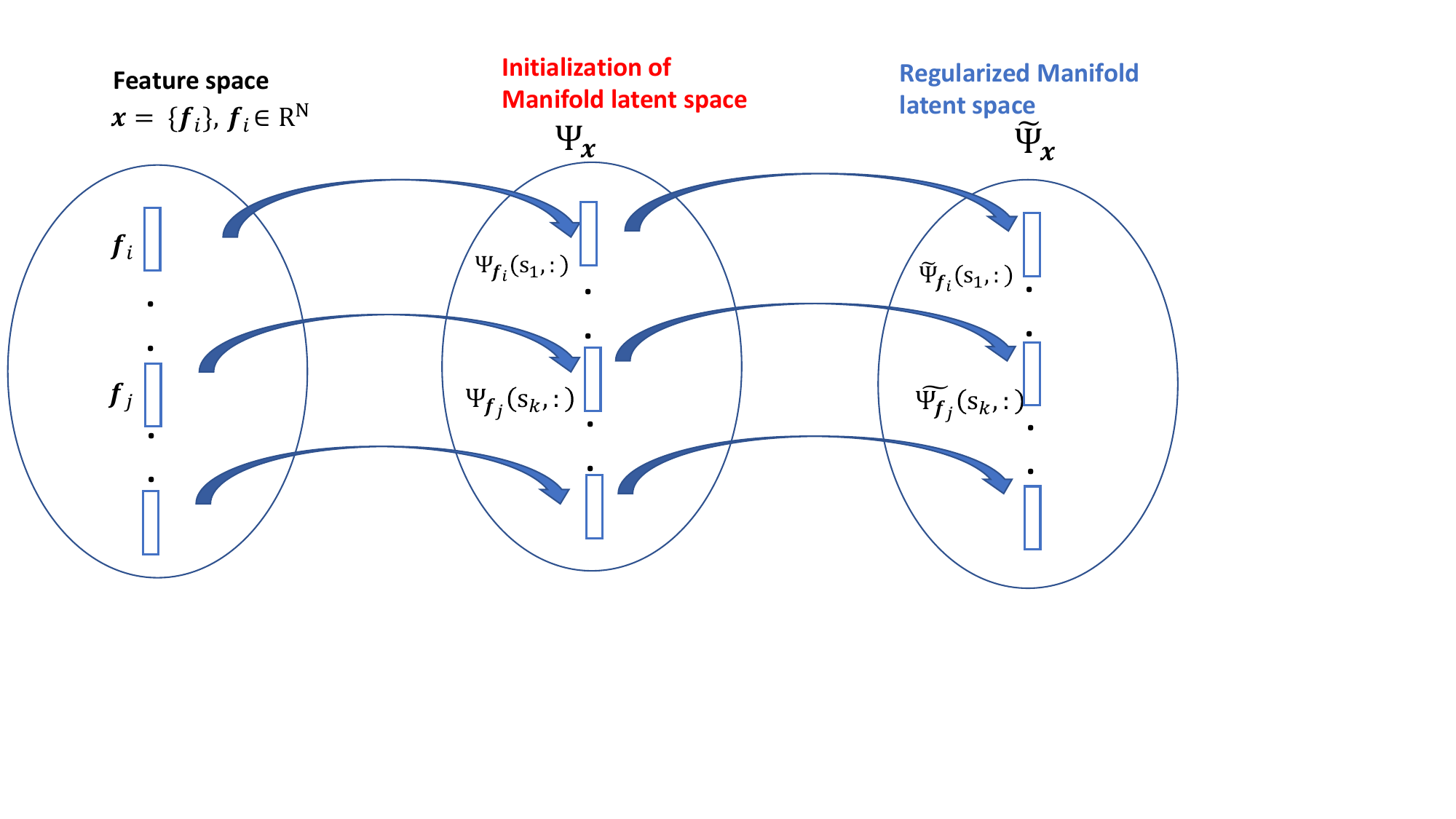}
\caption{The encoding step of our framework is illustrated as a mapping between each coordinate of the input features and a corresponding dimension in the embedding space. This mapping is facilitated through our proposed approach, which aligns the input dimensions of the features with the latent embedding space. By establishing this correspondence, we acquire the capability to interpret each dimension in the embedding space. Such interpretation allows for various analyses, including providing feature importance relative to the latent space.}
\label{fig:Mapping_illustration}
\end{figure}

\subsection{Additional details: Embedding Feature Importance using Mutual Information and Laplacian Score}
\label{sec:dditional details: Embedding Feature Importance using Mutual Information and Laplacian Score}

\subsubsection{Computing the Laplacian Score with respect to the embedding features }
The Laplacian score is calculated with respect to embedding features \( \tilde{\psi}_{\mathbf{x}} \) using the Laplacian graph \( \mathbf{L} \) and the degree matrix \( \mathbf{D} \).  To compute the Laplacian Score, we first subtract the mean and then calculate it as follows:
$ L_{s}(\tilde{\psi}_{\mathbf{x}})_l = \frac{(\tilde{\psi}_{\mathbf{x}})_l^{T} \mathbf{L}(\tilde{\psi}_{\mathbf{x}})_l}{(\tilde{\psi}_{\mathbf{x}})_l^{T} \mathbf{D}(\tilde{\psi}_{\mathbf{x}})_l}$. Smaller scores indicate that the embedding feature $(\tilde{\psi}_{\mathbf{x}})_l$ has a greater projection onto the subspace of eigenvectors corresponding to the smallest eigenvalues, signifying higher importance concerning the global graph structure. Therefore, the feature importance of $(\tilde{\psi}_{\mathbf{x}})_l$ can be directly interpreted as the importance of the corresponding original feature. Although the graph is typically constructed using local neighborhoods (e.g., via k-nearest neighbors), the Laplacian operator reflects global smoothness over the entire data manifold. 

\subsubsection{Embedding Feature Importance using Mutual Information}

We summarize the MI-based unsupervised feature selection procedure in Algorithm  \ref{alg:MI-based}., as described in Section  \ref{sec:Interpretable Embedding}. In comparison, applying the Laplacian Score(\cite{Ls}) to the embedding may be less effective for feature importance, as the embedding process can mix original features in ways that obscure their individual contributions. While the Laplacian Score prioritizes local neighborhood preservation, this does not necessarily correspond to discriminative power for clustering. In contrast, the proposed MI-based approach directly identifies features that are informative for distinguishing clusters. However, it relies on the quality of the pseudo-labels derived from clustering, which may not always reflect true cluster structure.

\begin{figure}[ht]
  \centering

  \begin{subfigure}[b]{0.45\columnwidth}
    \centering
    \includegraphics[width=\linewidth]{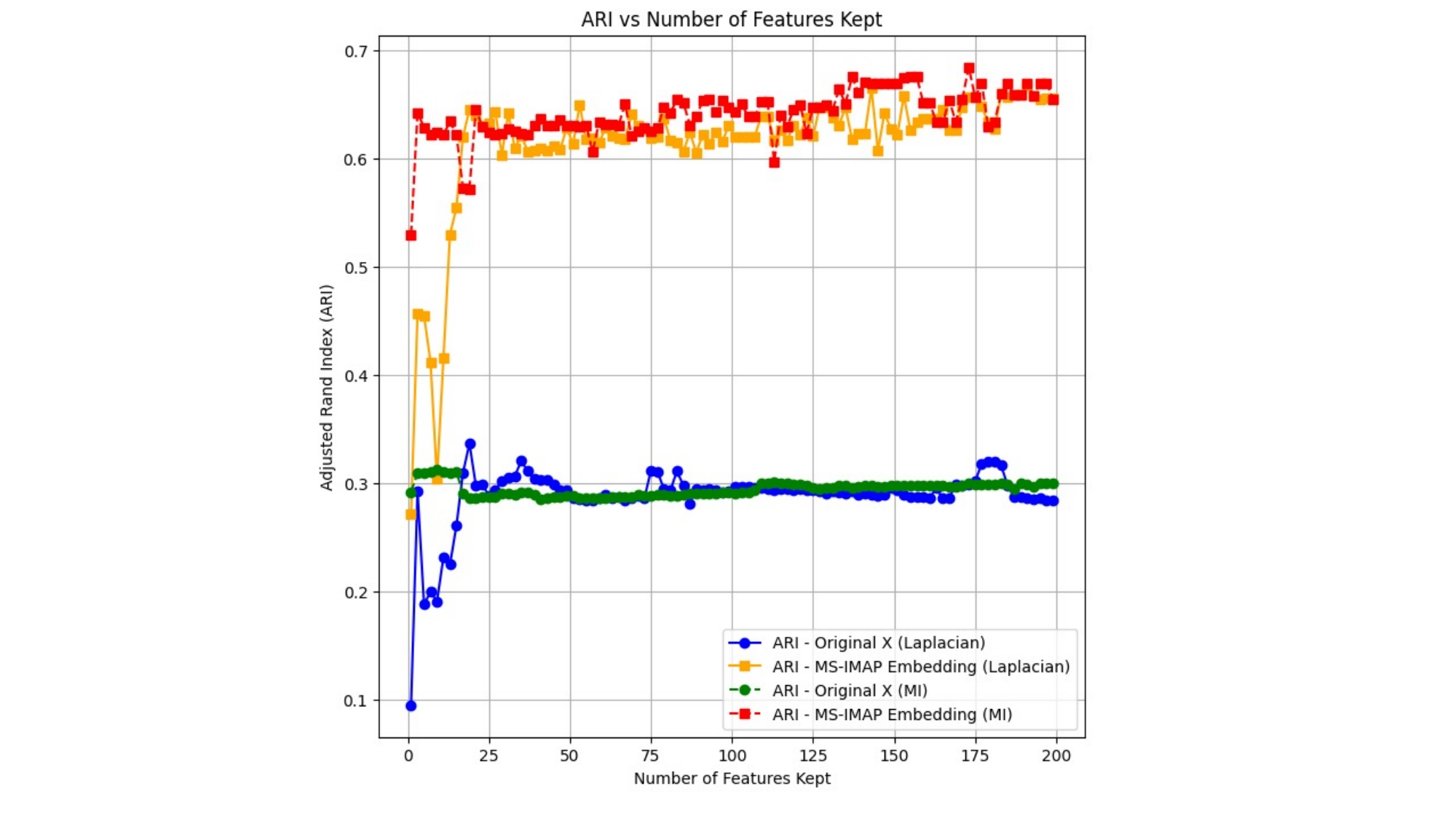}
    \caption{HAR dataset (ARI)}
    \label{fig:har_ari}
  \end{subfigure}
  \hfill
  \begin{subfigure}[b]{0.45\columnwidth}
    \centering
    \includegraphics[width=\linewidth]{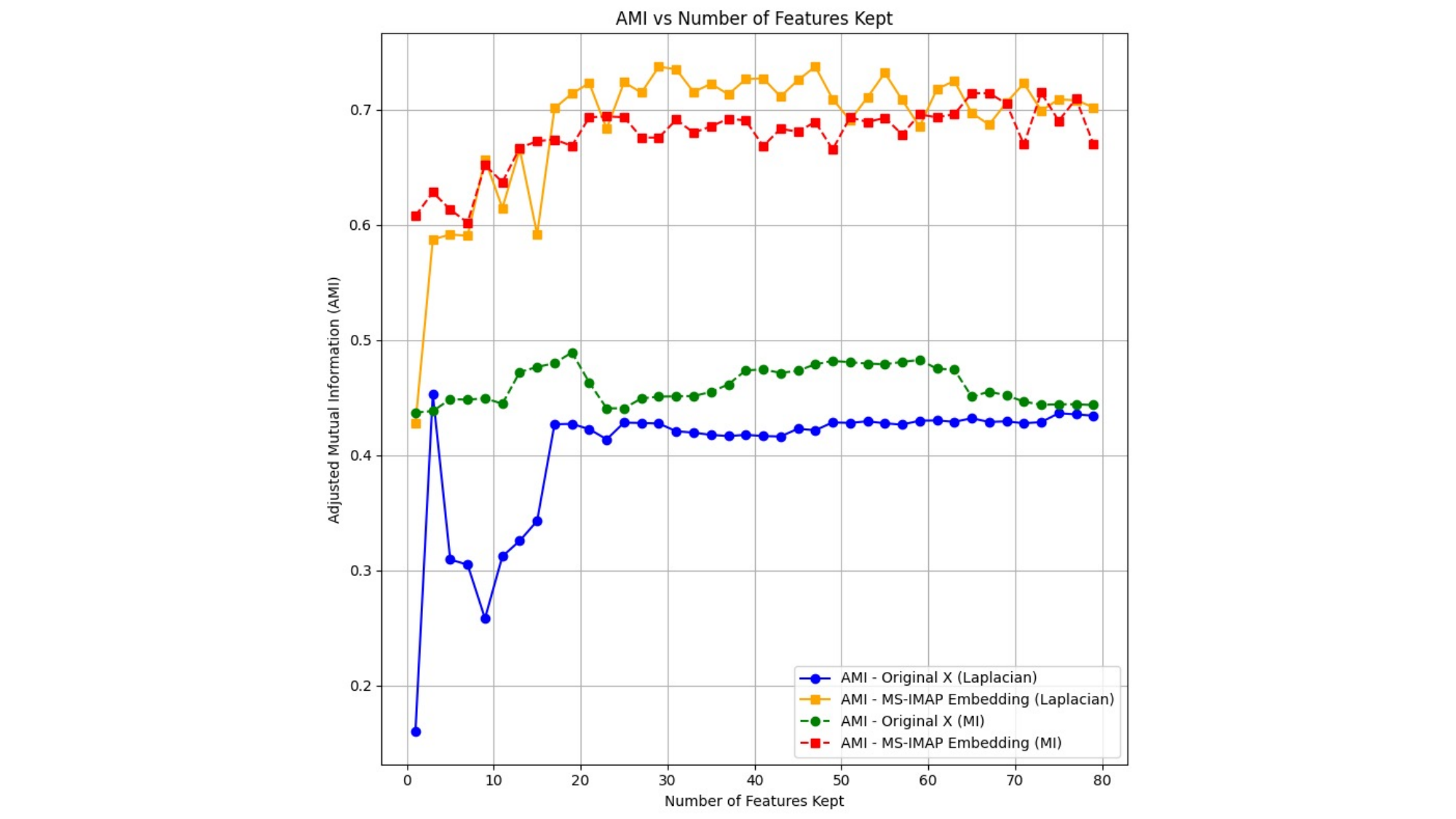}
    \caption{HAR dataset (AMI)}
    \label{fig:har_ami}
  \end{subfigure}

  \caption{Clustering results (ARI and AMI) on the HAR dataset using a subset of features based on feature importance, comparing MS-IMAP embedding space and the input features.}
  \label{fig:har_results}
\end{figure}

\begin{figure}[ht]
  \centering

  \begin{subfigure}[b]{0.45\columnwidth}
    \centering
    \includegraphics[width=\linewidth]{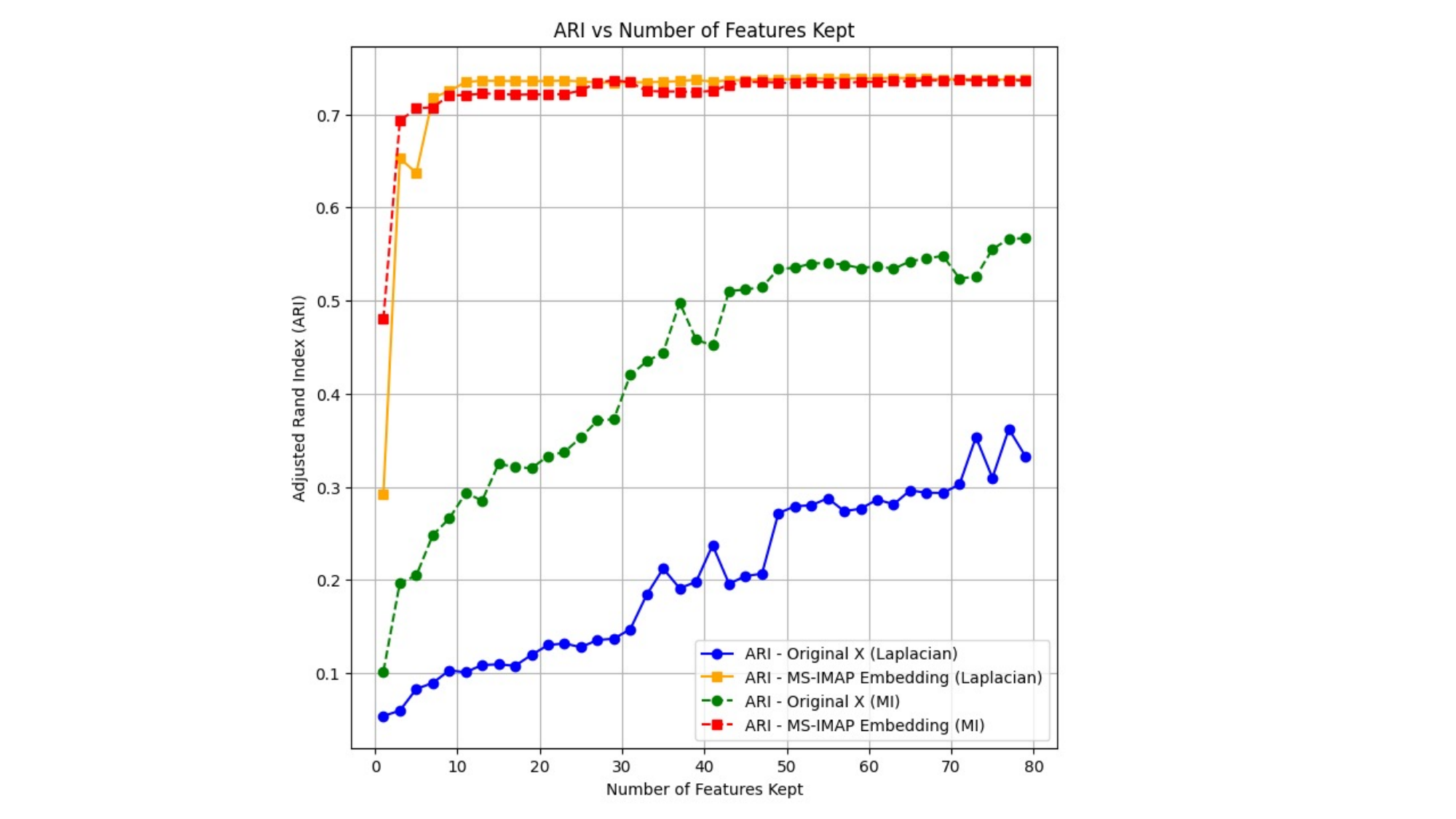}
    \caption{AWA dataset (ARI)}
  \end{subfigure}
  \hfill
  \begin{subfigure}[b]{0.45\columnwidth}
    \centering
    \includegraphics[width=\linewidth]{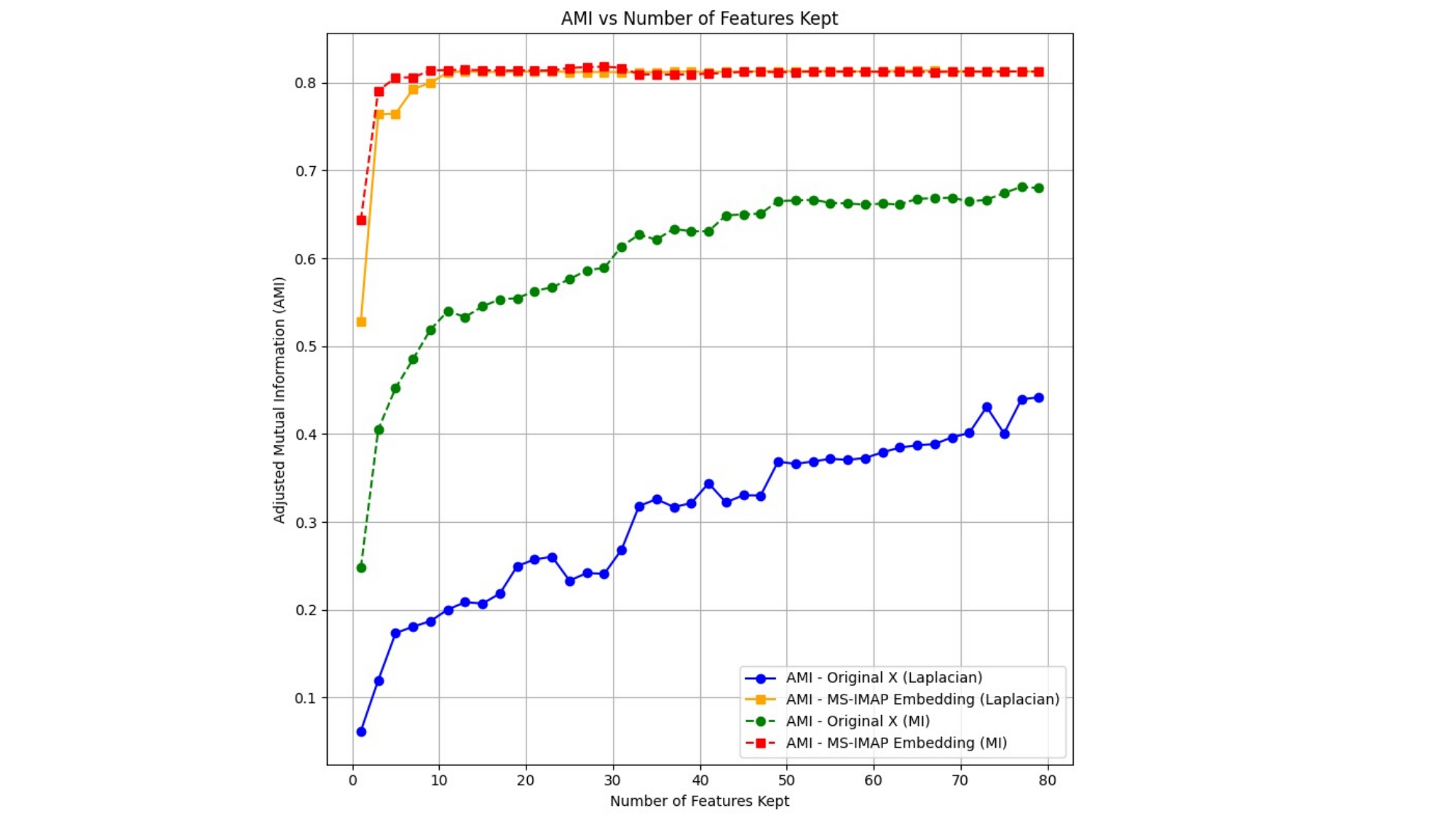}
    \caption{AWA dataset (AMI)}
  \end{subfigure}
\caption{Clustering results ARI and AMI on the AWA dataset using a subset of features based on feature importance, comparing MS-IMAP embedding space and the input features.}
  \label{fig:feat_import_AWA}
\end{figure}

We also include additional experiments that demonstrate the effectiveness of MS-IMAP for feature importance by evaluating clustering performance as a function of the number of selected features—complementing the results presented in Section \ref{sec:Interpretable Embedding}. Specifically, we assess the most significant features identified using Laplacian Score and Mutual Information-based selection, applied to both the MS-IMAP embedding and the original feature space.

Clustering results, evaluated using Adjusted Rand Index (ARI) and Adjusted Mutual Information (AMI), are obtained by applying $k$-means to different subsets of embedding features selected by each method, and compared to the corresponding subsets selected directly from the raw features.

On the HAR and AWA datasets, Figures \ref{fig:feat_import_HAR}, \ref{fig:feat_import_AWA}, show that, for both feature selection methods, clustering accuracy is consistently higher when using the MS-IMAP embedding than when using the original features—across a range of feature subset sizes and for both evaluation metrics. This improvement highlights the robustness of the MS-IMAP embedding, while maintaining interpretability through its one-to-one correspondence between embedding dimensions and original features.

\subsection{Dataset Details}
\label{sec:dataset-details}

\begin{figure}[t]
\centering
\includegraphics[width=0.9\columnwidth]{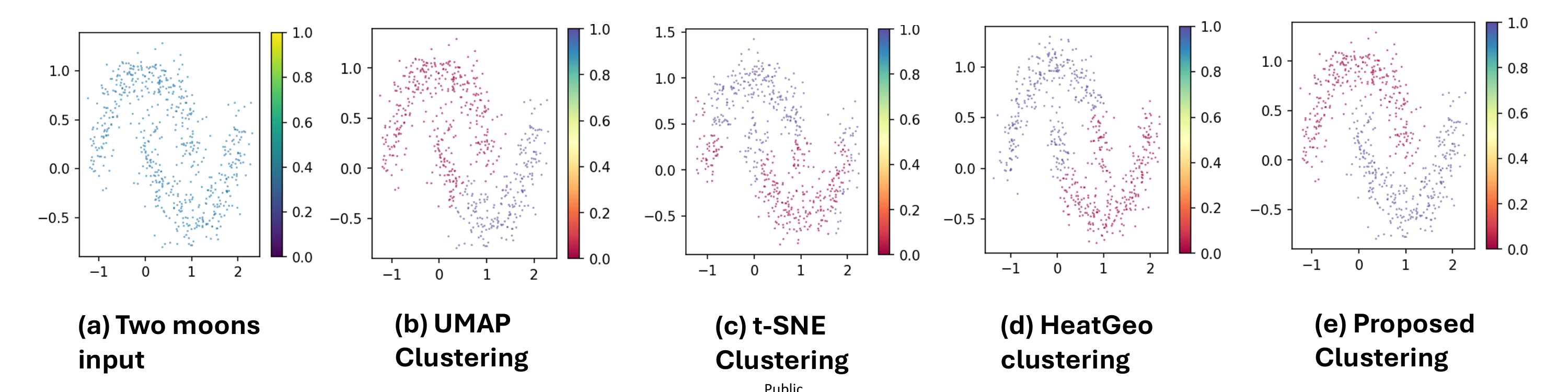}
\caption{Clustering results using two moons}
\label{fig:two_moons_cluster}
\end{figure}

\textbf{Two Moons.} The two moons dataset depicts two interleaving half-circles. We sampled $N=600$ points and used a Gaussian noise level having standard deviation $0.12$. An example is show in Figure \ref{fig:two_moons_cluster}. More specifically, to produce these points we use Sci-kit Learn's \texttt{sklearn.datasets.make\_moons} function with $n\_samples=600$ and $noise=0.12$. 

In Figure \ref{fig:two_moons_cluster}, we show example clusterings for each of the methods mentioned in the Experimental Results section of the main paper. \\
\textbf{Census.} From the UCI Machine Learning Repository \cite{UCI}, this dataset contains 14 features that are a mix of categorical, numerical, and binary. Such features include age, marital status, sex, etc. The goal is to predict whether a sample makes less than or equal to $\$50,000$, or strictly more. We use $32,561$ samples in our dataset. \\
\textbf{Lung Cancer.} The Zilionis dataset is widely used, and consists of single-cell RNA sequencing data. It has $306$ features, and $48,969$ samples. The data has 20 classes corresponding to cell type. More can be found at \cite{Zilionis}\\
\textbf{Animals with Attributes.} The Animals with Attributes (AWA) dataset, contains 5,000 data points corresponding to 10 unseen classes, where the testing image features are obtained from the pre-trained ResNet architecture, with $D= 2,048$ dimensions, and the semantic features are provided with $D = 85$ dimensions. More information can be found in Section 4.1 of \cite{Xian2017ZeroShotL}. \\
\textbf{Cora.} The Cora dataset has 2,708 scientific publications categorized into seven classes. The network has 5,429 links and each publication is described as a binary word vector indicating the presence or absence of a word. The dictionary consists of 1,433 words. In our experiments, we do not use the given graph, and only use the features. \cite{sen2008collective, mccallum2000automating}. \\
\textbf{HAR.} The Human Activity Recognition (HAR) dataset consists of sensor data collected from 30 subjects performing six different activities: walking, walking upstairs, walking downstairs, sitting, standing, and lying. The data was recorded using accelerometer and gyroscope sensors embedded in a smartphone worn on the waist. Each activity is represented as a feature vector of 561 attributes and 7352 samples, extracted from the raw sensor signals through signal processing techniques. In our experiments, we do not use the temporal dependencies and only utilize the extracted features.

\subsection{Experimental Results on two moons Synthetic Dataset}
\label{sec:more-experiments}
Here we show further experiments on more data, where we assess the robustness of our method using the two moons dataset, which is a 2D manifold depicting two interleaving half-circles. We sampled $N=600$ points, and set the Gaussian noise standard deviation to $0.12$. While spectral based methods such as UMAP are effective under relatively ``modest" noise levels, their performance deteriorates in the presence of larger amounts of noise. 
As shown in Table \ref{tab:ARI_AMI} under the \emph{Moons} column, our approach is robust and correctly clusters most points despite the large noise level, and competitive with the competing methods.

\begin{table}[htbp]
\centering
\begin{tabular}{lcccr}
\toprule
Data & \multicolumn{2}{c}{Two Moons dataset} \\
\midrule
Method/Accuracy & ARI & AMI \\    
\midrule
Features & 0.24 & 0.18  \\
UMAP &  0.54 &  0.51\\
t-SNE &  0.42 & 0.35 \\
ISOMAP & 0.36 & 0.3 \\    
Scattering Geometric.T & 0.26 & 0.19 \\ 
Diffusion Maps & 0.25 & 0.19 \\ 
PHATE & 0.42 & 0.35 \\
HeatGeo &0.54  & 0.52 \\
MS-IMAP  & \textbf{0.89} & \textbf{0.87} \\
\bottomrule
\end{tabular}
\caption{Comparison of clustering performance on the two moons dataset.}
\label{tab:dense-sparse}
\end{table}

\subsection{Hyperparameter tuning details}
\label{sec:hyperparameter-tuning-details}

We note that for UMAP, t-SNE, PHATE, PaCMAP, TriMap and HeatGeo, we hyperparameter tune the performance on each dataset, running the method five times for each hyperparameter configuration and taking the average. Then we report the best average score. 
Additional details regarding hyperparameter are provided below.  \\

We tune UMAP on each dataset in the Experimental Results section, by tuning over the parameter space in Table \ref{tab:umap-hyperparameters}.

\begin{table}[htbp]
    \centering
    \begin{tabular}{ll}
        \toprule
        \textbf{Hyperparameter} & \textbf{Set of values} \\ \midrule
        n\_neighbors & 2, 10, 15, 20, 30, 50, 100 \\ 
        min\_dist & 0, 0.1, 0.5, 0.99 \\ 
        n\_components & 2, 3 \\ \bottomrule
    \end{tabular}
    \caption{Space of parameters in which we tuned UMAP.}
    \label{tab:umap-hyperparameters}
\end{table}

\subsection{TSNE Hyperparameter Tuning Details}
\label{subsec:hyperparameter-tsne}

We tune TSNE on each dataset in the Experimental Results section, by tuning over the parameter space in Table \ref{tab:tsne-hyperparameters}.

\begin{table}[htbp]
    \centering
    \begin{tabular}{ll}
        \toprule
        \textbf{Hyperparameter} & \textbf{Set of values} \\ \midrule
        perplexity & 15, 30, 60, 200\\ 
        early\_exaggeration & 12, 24 \\ 
        n\_components & 2, 3 \\ \bottomrule
    \end{tabular}
    \caption{Space of parameters in which we tuned TSNE.}
    \label{tab:tsne-hyperparameters}
\end{table}

\subsection{PHATE Hyperparameter Tuning Details}
\label{subsec:hyperparameter-phate}

We tune PHATE on each dataset in the Experimental Results section, by tuning over the parameter space in Table \ref{tab:phate-hyperparameters}.

\begin{table}[htbp]
    \centering
    \begin{tabular}{ll}
        \toprule
        \textbf{Hyperparameter} & \textbf{Set of values} \\ \midrule
        knn & 3, 5, 10 \\ 
        decay (i.e. $\alpha$) & 6, 10, 14, 18, 40 \\ 
        t & auto, 50, 100, 150, 200 \\ \bottomrule
    \end{tabular}
    \caption{Space of parameters in which we tuned PHATE.}
    \label{tab:phate-hyperparameters}
\end{table}

\subsection{HeatGeo Hyperparameter Tuning Details}
\label{subsec:hyperparameter-heatgeo}

We tune HeatGeo on each dataset in the Experimental Results section, by tuning over the parameter space in Table \ref{tab:heatgeo-hyperparameters},

\begin{table}[htbp]
    \centering
    \begin{tabular}{ll}
        \toprule
        \textbf{Hyperparameter} & \textbf{Set of values} \\ \midrule
        knn & 5, 10, 15 \\ 
        lap\_type & normalized, combinatorial \\ 
        harnack\_regul & 0, 0.5, 1 \\ \bottomrule
    \end{tabular}
    \caption{Space of parameters in which we tuned HeatGeo.}
    \label{tab:heatgeo-hyperparameters}
\end{table}

\subsection{Ablation Studies of Hyperparameters for MS-IMAP}

Here we study what effect the hyperparameters -- the number of nearest neighbors, and the number of filters -- have on the clustering performance of MS-IMAP with method 2. In table \ref{tab:num-neighbors}, we demonstrate the effect of varying the number of neighbors, between 10, 15, and 20 neighbors. We see that the results are mostly the same, thus showing MS-IMAP is robust on real datasets.

\begin{table*}[htbp]
\centering
\begin{tabular}{lllllll}
\toprule
Dataset                        & \multicolumn{2}{l}{Census} & \multicolumn{2}{l}{Zilionis} & \multicolumn{2}{l}{AwA} \\ \midrule
Number of Neighbors / Accuracy & ARI          & AMI         & ARI           & AMI          & ARI        & AMI        \\ \midrule
10                             & 0.22         & 0.15        & 0.71          & 0.78         & 0.73       & 0.80       \\
15                             & 0.23& 0.15        & 0.70          & 0.76         & 0.74       & 0.81       \\
20                             & 0.22         & 0.15        & 0.70          & 0.77         & 0.71       & 0.79       \\ \bottomrule
\end{tabular}
\caption{Ablation study on MS-IMAP with varying the number of neighbors. The number of filters is kept at 5.}
\label{tab:num-neighbors}
\end{table*}

We also study the effect of varying the number filters. In table \ref{tab:num-filters}, we also see similar results when using different filters, showing the stability of MS-IMAP.

\begin{table*}[htbp]
\centering
\begin{tabular}{lllllll}
\toprule
Dataset                      & \multicolumn{2}{l}{Census} & \multicolumn{2}{l}{Zilionis} & \multicolumn{2}{l}{AwA} \\ \midrule
Number of Filters / Accuracy & ARI          & AMI         & ARI           & AMI          & ARI        & AMI        \\ \midrule
5                            & 0.23& 0.15        & 0.70          & 0.76         & 0.74       & 0.81       \\
6                            & 0.22         & 0.15        & 0.70          & 0.76         & 0.73       & 0.81       \\
7                            & 0.22         & 0.15        & 0.72          & 0.78         & 0.74       & 0.81       \\ \bottomrule
\end{tabular}
\caption{Ablation study on MS-IMAP with encoding method 2, varying the number of filters. The number of neighbors is kept at 15.}
\label{tab:num-filters}
\end{table*}

\subsection{Ablation Studies of Hyperparameters for t-SNE, Isomaps}

We study the effect of varying the number of neighbors for the methods: t-SNE, Isomaps, and Diffusion Maps. For t-SNE this would be the perplexity hyperparameter, for Isomaps this would be the number of neighbors, and for diffusion maps, this would be the parameter that affects the width of the Gaussian kernel, i.e. $\exp(\cdot / \alpha)$.

In Table \ref{tab:tsne-perplexity}, we see choosing a smaller perplexity of 15 does worse than the perplexity of 30, 60 and 200.

\begin{table*}[htbp]
\centering
\begin{tabular}{lllllll}
\toprule
Dataset               & \multicolumn{2}{l}{Census} & \multicolumn{2}{l}{Zilionis} & \multicolumn{2}{l}{AwA} \\ \midrule
Perplexity / Accuracy & ARI          & AMI         & ARI           & AMI          & ARI        & AMI        \\ \midrule
15                    & 0.03         & 0.04        & 0.37          & 0.67         & 0.69       & 0.76       \\ 
30                    & 0.15         & 0.15        & 0.38          & 0.68         & 0.73       & 0.80       \\ 
60                    & 0.17         & 0.17        & 0.38          & 0.69         & 0.73       & 0.80       \\ 
200                   & 0.2        & 0.24        & 0.39          & 0.69         & 0.71       & 0.80       \\
\bottomrule
\end{tabular}
\caption{Ablation study on t-SNE, varying the perplexity.}
\label{tab:tsne-perplexity}
\end{table*}

In Table \ref{tab:isomap-neighbors}, we see a similar pattern to t-SNE, where choosing too low causes reduction in performance. But performance stabilizes around choosing the number of neighbors as 5-10+.

\begin{table*}[htbp]
\centering
\begin{tabular}{lllllll}
\toprule
Dataset              & \multicolumn{2}{l}{Census} & \multicolumn{2}{l}{Zilionis} & \multicolumn{2}{l}{AwA} \\ \midrule
Neighbors / Accuracy & ARI          & AMI         & ARI           & AMI          & ARI        & AMI        \\ \midrule
2                    & -0.05        & 0.02        & 0.41          & 0.55         & 0.43       & 0.58       \\ 
5                    & 0.18         & 0.09        & 0.44          & 0.56         & 0.52       & 0.60       \\ 
10                   & 0.18         & 0.09        & 0.44          & 0.57         & 0.51       & 0.62       \\ \bottomrule
\end{tabular}
\caption{Ablation study on Isomap, varying the number of neighbors.}
\label{tab:isomap-neighbors}
\end{table*}

\newpage

\subsection{Fast computation using Chebyshev polynomials}

We provide additional details regarding the fast computation of SGW coefficients \cite{Hammond}.  Directly computing the SGW coefficients above requires calculating the entire eigensystem of the Laplacian, which is computationally intensive - $O(N^{3})$ for \(N\) points. Instead, Hammond et al. \cite{Hammond} suggested computing the SGW using a fast algorithm based on approximating the scaled generating kernels through low-order polynomials. The wavelet coefficients at each scale are then computed as a polynomial of \(\mathbf{L}\) applied to the input data, using approximating polynomials given by truncated Chebyshev polynomials.

The Chebyshev polynomials \(T_{k}(y)\) are computed using the recursive relations:
$T_{k}(y) = 2yT_{k-1}(y) - T_{k-2}(y)$  for  \(k \geq 2\), where \(T_{0} = 1\) and \(T_{1} = y\).

The SGW coefficients are then approximated using wavelet and scaling function coefficients as follows:
\begin{equation}
\psi_{f}(s_{j},i) \sim  \left (  \frac{1}{2}c_{j,0}f + \sum_{k=1}^{K}c_{j,k}\bar{T}_{j,k}(\mathbf{L})f  \right )_{i}
\end{equation} 

where $c_{j,k}, j > 0$ are the Chebyshev coefficients and \(\bar{T}_{j,k}\) are the shifted Chebyshev polynomials \(\bar{T}_{k}(x) = T_{k}\left(\frac{x-a}{a}\right)\) for \(x \in [0,\lambda_{\mbox{max}}]\), where $x = a(y + 1)$, a = $\lambda_{ \mbox{max} }$. 
The scaling function coefficients, which are corresponding to a low-pass filter operation, are approximated in a similar way using Chebyshev polynomials. Note that the scaling kernel function is a low pass filter $h$ satisfying $h(0)>0$  and $h(x) \rightarrow 0$ when $ x\rightarrow \infty$. If the graph is sparse, we obtain a fast computation of the matrix-vector multiplication $\bar{T}_{j,k}(\mathbf{L})f$, where the computational complexity scales linearly with the number of points, resulting in a complexity of $O(N)$ for an input signal $f \in \mathbb{R}^{N}$. The SGWs efficiently map an input graph signal (a vector of dimension $N$) to $NK$ scaling and wavelet coefficients. \\

\subsection{Runtime and Computational Complexity}

We have performed experimental runtime studies on empirical datasets. The execution time of our method with Python code implementation with 32 cores Intel Xeon 8259CL running at 2.50Ghz and 256GB of RAM on the Cancer QC data-set of 48,969 samples and 306 features took 27 mins, on Cora 5 min, on the AWA dataset 1.35 mins and on Census 7 min.  In fields such as finance and genomics/biology, where typical dataset sizes range from under 100,000 to several million data points, analysis is often conducted offline once the graph embedding is computed. Runtimes of up to a few hours are generally acceptable and do not present a significant challenge, making the method well-suited for real-world applications. 

The computational complexity of Multi-Scale UMAP is of $O(ND log(N))$ for construction of the multi-scale representations which includes the $k$ nearest neighbor graph using k-d tree, the SGW transform which is $O(N)$ for each dimension of the manifold for sparse graphs. The optimization stage has a complexity which scale with the number of edges in the graph, which has a complexity of $O(kDN)$. Note that our approach generally requires the embedding dimensionality to be at least equal to the input dimensionality of the features, which adds computational complexity. 

\newpage

\subsection{Theoretical results: Sampling set for Smooth Manifolds with functions defined over Paley-Wiener Spaces} 
\label{sec:Theoreticalresultsproofs}

In this section, we characterize the theoretical properties of the representation power of the SGW operator by considering functions sampled from the Paley-Wiener spaces \cite{Pesenson2008SamplingIP} on combinatorial graphs. The Paley-Wiener spaces were introduced on combinatorial graphs in \cite{Pesenson2008SamplingIP} and a corresponding sampling theory was developed which resembles the classical one. Pesenson proved in \cite{Pesenson2008SamplingIP} that Paley-Wiener functions of low type are uniquely determined by their values on certain subgraphs (which are composed from a set of nodes known as the uniqueness sets) and can be reconstructed from such sets in a stable way.
We demonstrate that the SGW operator can represent functions $f$ that reside in the Paley-Wiener space on combinatorial graphs more efficiently than the graph Laplacian operator \( \mathcal{L} \). The effectiveness of the SGW operator representation in this case can be understood in several ways. In one way, by the ability of the SGW operator to accurately represent functions with larger bandwidth, i.e., $f \in PW_{\omega'}(G)$ where $\omega < \omega' $. 
We first summarize the main notions and definitions. 
The space $L_{2}(G)$ is the Hilbert space of all complex-valued functions $f: V \rightarrow \mathbb{C} $ with the following inner product  $\left \langle f,g \right \rangle = \sum_{v \in G} f(v)\overline{g(v)}$ and the norm 

\begin{equation}
    ||f|| =  \left (  \sum_{v \in V} {|f(v)|^{2}} \right )^{1/2}   
\end{equation}
The Laplace (normalized) operator $\cal L$  is defined by the formula \cite{Pesenson2008SamplingIP}: 

\begin{equation}
  \mathcal{L} f(v) =   \frac{1}{\sqrt{d(v)}} \sum_{u \sim v} \left ( \frac{f(v)}{\sqrt{d(v)}} - \frac{f(u)}{\sqrt{d(u)}} \right ), f\in L_{2}(G)
\end{equation}

In order to prove these results, we will first need the following definitions: 
\begin{definition}
The Paley-Wiener space of $\omega$ -bandlimited signals $f\in L_2(G)$ is defined as follows: 

\begin{equation}
    PW_{\omega}(G) = \left \{f|\hat{f}(\lambda)= 0 \,\, \forall \,\, \lambda> \omega )  \right \}
\end{equation}

\end{definition}

We consider a simple, undirected, unweighted, and connected graph $G = (V, \mathbf{W})$, where $V$ is its set of $N$ vertices and $\mathbf{W}$ is its set of edges. The degree of $v$ is number of vertices adjacent to a vertex $v$ is and is denoted by $d(v)$. We assume that degrees of all vertices are bounded by the maximum degree denoted as 
\begin{equation}
    d(G) = \mbox{max}_{v\in V} d(v)
\end{equation}

The following definition \cite{Pesenson2008SamplingIP} explains the uniqueness set.

\begin{definition} A set of vertices $U \subset   
 V$ is a uniqueness set for a space $PW_{\omega}(G)$ if for every two functions from $PW_{\omega}(G)$ that coincide on $U$, then they coincide on $V$. 
\end{definition}

\begin{definition}
For a subset $S \subset   
 V$, denote $L_{2}(S)$ as the space of all functions from $L_{2}(G)$ with support in $S$: $$ L_{2}(S) = \left \{ \varphi \in L_{2}(G), \varphi(v) = 0, v \in V(G)\setminus S  \right \}  $$
\end{definition}

\begin{definition}\cite{Pesenson2008SamplingIP}
We say that a set of vertices $S \subset   
 V$ is a $\Lambda$ -set if for any $\varphi \in L_{2}(S)$ it admits a Poincare inequality with a constant $\Lambda = \Lambda (S)>0$  
 \begin{equation}
     ||\varphi || \leq \Lambda ||\mathbf{\cal L}\varphi ||, \, \varphi  \in L_{2}(S)
 \end{equation}
The infimum of all $\Lambda(S) > 0$ for which $S$  is a $\Lambda$ -set will be called the Poincare constant of the set $S$ and denoted by $\Lambda$.  
\end{definition} 
The definition \label{Removable_set} above provides a tool to determine when bandlimited signals in Paley-Wiener spaces $PW_{\omega}(G)$ can be uniquely represented from their samples on a given set. The role of $\Lambda$-sets was explained and proved in the following Theorem by Pesenson \cite{Pesenson2008SamplingIP}, that shows that if $S \subset V$ then any signal $f \in PW_{\omega}(G)$ can be uniquely represented by its function values in the complement set $U = V(G) \setminus S$:
\begin{Theorem} \cite{Pesenson2008SamplingIP}  
\label{L1}
If $S \subset V$ is a $\Lambda $- set, then the set $U = V(G) \setminus S$ is a uniqueness set for any space $PW_{\omega}(G)$ with $\omega <  \frac{1}{\Lambda}$.  
\end{Theorem}

\textbf{Remark:} Note that non-trivial uniqueness sets can not exist for functions from any Paley-Wiener subspace $PW_{\omega}(G)$ with any $\lambda_{0}\leq \omega \leq \lambda_{N}$, but they can exist for some range $\lambda_{0} \leq\omega <\Omega$, as was shown in \cite{Pesenson2008SamplingIP}. \\

We state one of our main results, in which we employ the SGW operator \( \psi \) to characterize the uniqueness set using the \( \Lambda_{\psi} \)-set, therefore extending the \( \Lambda \)-set concerning the graph Laplacian operator $\cal L$. 
\begin{Theorem}
\label{SGW_bound}
Let $G= (V, \mathbf{W})$ be a connected graph with $N$ vertices. Assume that there exist a set of vertices $S \subset V$ for which the conditions (1)-(2) in Lemma \ref{L3} below hold true. Let $\psi$ be the SGW operator using a polynomial $p(\mathbf{\cal L})$ with the coefficients $\left \{  a_{k} \right \}_{k=0}^{K}$ such that $\psi_{f} = \sum_{k=0}^{K} a_{k}\mathbf{\cal L}^{k}f$. Then, for any $\varphi \in L_2({S})$, we have that the following inequality holds: 
\begin{equation}
||\varphi|| \leq \Lambda_{\psi_{ }}  || \psi_{ {\varphi}}||  
\end{equation}
and thus the set $S$ is a $\Lambda_{\psi}$-set for the operator $\psi$ with $\Lambda_{\psi_{ }} = {\frac{1}{ \sqrt { \sum_{k=0}^{K} \frac{a_{k}^{2}}{\Lambda^{2k}} }} } $. 
\end{Theorem}

We recall the following results from \cite{Pesenson2008SamplingIP}. Note that \cite{Pesenson2008SamplingIP} established the construction of a \( \Lambda \)-set by imposing specific assumptions on the sets \(S\) and \(U\). Our result Theorem \ref{SGW_bound} holds similar assumptions as in \cite{Pesenson2008SamplingIP}.

\begin{Lemma}[Lemma 3.6 of \cite{Pesenson2008SamplingIP}]
\label{L3} Given a connected graph $G = (V, \mathbf{W})$, a set of vertices $ S \subset V $, its complement $U = V \setminus S $, for which the following conditions hold true

\begin{enumerate}
    \item For every $s \in S$ there exists $u \in U$ that is a neighbor of $s$, i.e., $w(u,s)>0$.
    \item For every $s \in S$ there exists at least one neighbor node $u \in U$ whose adjacency set intersects $S$ only over $s$. 
\end{enumerate}

Then there exist a set of vertices $S \subset V$ which is a $\Lambda$-set, with $ \Lambda = d(G)$. 
\end{Lemma}

We require one more property of the Laplacian operation, Lemma \ref{laplacepowers} below. To this end, we use a lemma from \cite{Pesenson2008SamplingIP},

\begin{Lemma}[Lemma 3.9 of \cite{Pesenson2008SamplingIP}]
\label{lemma3.9Pesenson}
    If $S$ is a $\Lambda$-set, then for any $\varphi \in L_2(S)$ and all $t \ge 0$, $k = 2^l$, $l=0, 1, 2, ...$
    \begin{equation*}
        \frac{1}{\Lambda^k} \|\mathcal{L}^t \varphi\| \le \|\mathcal{L}^{k+t} \varphi\|, \quad \varphi \in L_2(S)
    \end{equation*}
\end{Lemma}

Now we are ready to prove a modification of Lemma 3.9 from \cite{Pesenson2008SamplingIP}:

\begin{Lemma}
    \label{laplacepowers}
    Let $S$ be a $\Lambda$-set, $\varphi \in L_2(S)$, and let $m \ge 1$ be an integer. Then
        \begin{equation*}
            \frac{1}{\Lambda^m} \|\varphi\| \le \|\mathcal{L}^m \varphi\|.
        \end{equation*}
\end{Lemma}
\textbf{Proof:}
    Because every positive integer has binary representation, then we have for some $j_i\in \{0, 1, 2, ...\}, i=1, 2, ..., N$, that
        \begin{equation*}
            m = \sum_{i=1}^N 2^{j_i}.
        \end{equation*}
    Using Lemma \ref{lemma3.9Pesenson}, let $k=2^{j_1}$ and $t = 2^{j_2} + \cdots + 2^{j_N}$, so $m = k + t$. Then we have
        \begin{align*}
            \|\mathcal{L}^{m}\varphi\| &= \|\mathcal{L}^{k + t}\varphi\| \\
            &\ge \frac{1}{\Lambda^k} \|\mathcal{L}^{t} \varphi\| \\
            &= \frac{1}{\Lambda^{2^{j_1}}} \|\mathcal{L}^{2^{j_2} + \cdots + 2^{j_N}} \varphi\|
        \end{align*}
    We can now let $k_2 = 2^{j_2}$ and $t_2 = 2^{j_3} + \cdots + 2^{j_N}$ so that,
        \begin{align*}
            \|\mathcal{L}^{m}\varphi\| \ge \frac{1}{\Lambda^{2^{j_1}}} \|\mathcal{L}^{2^{j_2} + \cdots + 2^{j_N}}\varphi\| &= \frac{1}{\Lambda^{2^{j_1}}} \|\mathcal{L}^{k_2 + t_2} \varphi\| \\
            & \ge \frac{1}{\Lambda^{2^{j_1}}} \frac{1}{\Lambda^{k_2}} \|\mathcal{L}^{t_2} \varphi\| \\
            &= \frac{1}{\Lambda^{2^{j_1} + 2^{j_2}}} \|\mathcal{L}^{2^{j_3} + \cdots + 2^{j_N}} \varphi\|
        \end{align*}
    Continuing in this way, we eventually get,
        \begin{align*}
            \|\mathcal{L}^m\varphi\| \ge \frac{1}{\Lambda^m} \|\varphi\|.
        \end{align*}
    completing the proof. $\square$

In the next theorem, we expand the characterization of the uniqueness set using the \( \Lambda \)-set concerning the graph Laplacian operator $\mathcal{L}$ to include cases where we employ the SGW operator \( \psi \). We thereby characterize the uniqueness set using the \( \Lambda_{\psi} \)-set for the SGW operator. \\
We now turn to prove Theorem \ref{SGW_bound}, which was stated earlier.  \\
\textbf{Proof of Theorem \ref{SGW_bound}:}\\
Let $S \subset V$ be a set satisfying the conditions in Lemma \ref{L3}, so $S$ is $\Lambda$-set. Then by Lemma \ref{laplacepowers}, for any integer $k \ge 1$, 
\begin{equation*}
\frac{1}{\Lambda^k} \|\varphi\| \le \|\mathcal{L}^k \varphi\|.
\end{equation*}
The SGW operator of $\varphi$ has the form,
\begin{equation*}
    \psi_\varphi := \sum_{i, j} \sum_{k=0}^{K} a_{k}\mathbf{\cal L}^{k}\varphi  
\end{equation*}
for some $K\ge 1$, $a_k \in \mathbb{R}$ are coefficients, and where the summation over $i$ and $j$ represents the summation over the nodes of the graph. Now we calculate,
\begin{align*}
    \|\psi_\varphi\| &= \left( \sum_{i,j} \sum_{k=0}^K \left| a_k (\mathcal{L}^k \varphi)_{ij} \right|^2 \right)^{1/2} \\
    &= \left( \sum_{i,j} \sum_{k=0}^K a_k^2 \left|(\mathcal{L}^k \varphi)_{ij} \right|^2 \right)^{1/2} \\
    &= \left( \sum_{k=0}^K a_k^2 \sum_{i,j}  \left|(\mathcal{L}^k \varphi)_{ij} \right|^2 \right)^{1/2} \\
    &= \left( \sum_{k=0}^K a_k^2 \|\mathcal{L}^k \varphi\|^2 \right)^{1/2} \\
    &\ge \left( \sum_{k=0}^K a_k^2 \frac{1}{\Lambda^{2k}} \|\varphi\|^2 \right)^{1/2} \\
    & =  \|\varphi\| \left( \sum_{k=0}^K a_k^2 \frac{1}{\Lambda^{2k}} \right)^{1/2}
\end{align*}
Letting
\begin{equation*}
    \Lambda_\psi := \frac{1}{ \sqrt{ \sum_{k=0}^K \frac{a_k^2}{\Lambda^{2k}} } }
\end{equation*}
Then we have $\|\varphi\| \le \Lambda_\psi \|\psi_\varphi\|$, so $S$ is also a $\Lambda_\psi$-set.  $$\square$$

From the assumptions and proof of Theorem \ref{SGW_bound}, we get the immediate result:

\begin{Corollary}
    Let $S\subset V$ be a $\Lambda$-set. Then $S$ is also a $\Lambda_\psi$-set.
\end{Corollary}

\textbf{Remark 1:} An important property which can be observed from Theorem  \ref{SGW_bound} is the following: given the SGW $\psi_{ {f}}$ as a spectral representation operator applied on $f$, we can choose coefficients $\left \{  a_{k} \right \}_{k=0}^{K}$ such that we obtain a $\Lambda_{\psi}$ - set associated with the operator $\psi_{ }$, which is smaller than the $\Lambda$ - set associated with the Laplacian operator $\cal L$.
This implies that the operator $\psi_{ }$ provides more flexibility and better control over smoothness properties in comparison to the Laplacian operator.  \\
\textbf{Remark 2: } Since the $\Lambda_{\psi}$ - set can be chosen to be smaller than $\Lambda$ - set (for a proper choice of the coefficients $\left \{  a_{k} \right \}_{k=0}^{K}$ using the operator $\psi_{ }$) then the SGW  operator $\psi_{ }$ provides a more efficient representation 
for $f \in  PW_{\omega'}(G)$ with $\omega < \omega'$ using the same subsets of nodes from the uniqueness set $U$ in comparison to the Laplacian operator $\cal L$. \\
\textbf{Remark 3:}  Note that the characterization of the the uniqueness set does not rely on a reconstruction method of the graph signal values of $f(S)$ from their known values on $U$. \\
The next Theorem demonstrates the role of  the $\Lambda_{\psi_{ }}$ -set with respect to the operator $\psi_{ }$, where we show that any signal $f\in PW_{\omega}(G)$ can be uniquely represented by its samples on the uniqueness set $U$. This results resembles the role of $\Lambda$-sets with respect to the graph Laplacian operator $\mathbf{\cal L}$, yet with a different bound then Lemma \ref{L1}. 

\begin{Theorem}
\label{Wavelet_uninqueness}
Let $G= (V, \mathbf{W})$ be a connected graph with $N$ vertices and $f\in  PW_{\omega}(G)$ for $\lambda_{1} < \omega < \lambda_{\mbox{max}}$. Given the SGW operator $\psi$, and a set $S$ which is a $\Lambda_{\psi}$ - set. Then the set $U = V \setminus S$ is a uniqueness set for any space $PW_{\omega}(G)$ with $\omega < 1/\Lambda_{\psi}$. 
\end{Theorem}
\textbf{Proof:} Given $f,\tilde{f} \in  PW_{\omega}(G)$, then $f-\tilde{f} \in PW_{\omega}(G)$. Assume that $f \neq \tilde{f}$. 
If $f,\tilde{f} $ coincide on $U = V \setminus S$, then $f-\tilde{f} \in L_{2}(S)$ and therefore 

\begin{equation}
    || f-\tilde{f}|| \leq \Lambda_{\psi} ||\psi_{ f - \tilde{f}} || 
\end{equation}
Since $\psi_{ {f}} \in \mathbb{R}^{N}$, we have that by properties of a vector space in $\mathbb{R}^{N}$, using the Cauchy–Schwarz inequality and assuming $|a_{k}|\leq 1 \, \forall k$, we have:

\begin{equation}
 || \psi_{ f- \tilde{f}} || \leq \omega ||f- \tilde{f}||   
\end{equation}
Combining the inequalities above and using the inequality $\Lambda_{\psi}\omega < 1$ we have that 
\begin{equation}
        || f-\tilde{f}|| \leq \Lambda_{\psi}||\psi_{ f-\tilde{f}} ||  \leq  \Lambda_{\psi_{ }} \omega|| f-\tilde{f}|| <||f-\tilde{f}|| 
\end{equation}

which is a contradiction to the assumption that $f \neq \tilde{f}$. Thus, the set $U = V \setminus S $ is a uniqueness set for any space $PW_{\omega}(G)$ with $\omega < 1/\Lambda_{\psi}$.                                                     $\square$  \\\\
\textbf{Remark 1:} Note that $\Lambda\omega <1$ implies that $\Lambda_{\psi}\omega < 1$ given $\Lambda_{\psi} < \Lambda$, then we can increase the size number of nodes in the uniqueness set $U$ for $PW_{\omega}(G)$  (for a proper choice of the coefficients $\left \{  a_{k} \right \}_{k=0}^{K}$ using the operator $\psi_{ }$). In other words, we may increase the size of $S$ (thus reducing the size of $U$) and still obtain a uniqueness set with a smaller size for the graph signals in $PW_{\omega}(G)$.  \\
\textbf{Remark 2:} We note that the results of Theorem \ref{Wavelet_uninqueness} concerning the uniqueness set are independent from the stability properties of the representation. In order to achieve stability which is important for reconstruction, it is required to construct a wavelet operator using multiple scales $t_{j}, j=1,...J$, as proposed in \cite{Hammond}. More specifically, we can express the function $\varphi$ using multiple scales $t_{j}$ (here we replace the previous notation of scale $s_{j}$ with $t_{j}, j=1,..J$, not to confuse with nodes $s \in S $). Then for a fixed scale $t_{j}$ we have that the SGW is given by  
$|| \psi_{\varphi}(t_{j})|| = \left (  \sum_{i=1}^{N} \sum_{k=0}^{K} |a_{t_{j},k} ({\cal L}^{k}\varphi(i))|^{2} \right )^{1/2} $ 
and 
\begin{align}
    || \psi_{\varphi} ||_ = \left ( \sum_{j}\sum_{i=1}^{N} \sum_{k=0}^{K} |a_{t_{j},k} ({\cal L}^{k}\varphi(i))|^{2} \right )^{1/2}
\end{align}
In a similar way to the arguments provided in Theorem \ref{SGW_bound} we can choose coefficients $a_{t_{j},k}$ associated with the Laplacian polynomial such that the inequality  $ ||\varphi|| \leq \Lambda_{\psi_{ }}  || \psi_{ {\varphi}}||  $ is satisfied. \\
\textbf{Remark 3: } Optimal Choice of coefficients \( a_k \) for Minimizing $ \Lambda_{\psi} $ for optimal choice of $ \Lambda_{\psi} $ in the multi-scale setting.  \\
To find the coefficients \( a_{t_j, k} \) that minimize \( \Lambda_{\psi} \) where:
\begin{equation}
    \Lambda_{\psi} = \frac{1}{\sqrt{\sum_{j=1}^{J} \sum_{k=0}^{K} \frac{a_{t_j, k}^2}{\Lambda^{2k}}}}
\end{equation}

subject to the constraint: $
    \sum_{j=1}^{J} \sum_{k=0}^{K} a_{t_j, k}^2 = 1$, we optimize the coefficients $ a_{t_j, k} $, which control how much of the signal at scale \( t_j \) and power \( k \) is captured by the SGW operator.
We use the Lagrangian $  \mathcal{F}$, incorporating the energy constraint into the optimization problem and Define: \\

\begin{equation}
   \mathcal{F}(a_{t_1, 0}, \dots, a_{t_J, K}, \mu) = \frac{1}{\sqrt{\sum_{j=1}^{J} \sum_{k=0}^{K} \frac{a_{t_j, k}^2}{\Lambda^{2k}}}} + \mu \left( \sum_{j=1}^{J} \sum_{k=0}^{K} a_{t_j, k}^2 - 1 \right) 
\end{equation}

where$\mu$  is the Lagrange multiplier that ensures the total energy of the coefficients is equal to 1. \\ 
Taking partial derivatives, solving for $ a_{t_j, k} $, and differentiating \( \mathcal{L} \) with respect to each coefficient $a_{t_j, k}$:

\begin{equation}
    \frac{\partial \mathcal{F}}{\partial a_{t_j, k}} = \frac{- a_{t_j, k} / \Lambda^{2k}}{\left( \sum_{j=1}^{J} \sum_{k=0}^{K} \frac{a_{t_j, k}^2}{\Lambda^{2k}} \right)^{3/2}} + 2 \mu a_{t_j, k} = 0.
\end{equation}

rearranging the equation above we obtain:
\[
\frac{a_{t_j, k}}{\Lambda^{2k}} = 2\mu a_{t_j, k} \left( \sum_{j=1}^{J} \sum_{k=0}^{K} \frac{a_{t_j, k}^2}{\Lambda^{2k}} \right)^{3/2}
\]

assuming \( a_{t_j, k} \neq 0 \), we can divide both sides of the equation by \( a_{t_j, k} \), yielding: \\

\begin{equation}
\frac{1}{\Lambda^{2k}} = 2\mu \left( \sum_{j=1}^{J} \sum_{k=0}^{K} \frac{a_{t_j, k}^2}{\Lambda^{2k}} \right)^{3/2}    
\end{equation}

thus the sum \( \sum_{j=1}^{J} \sum_{k=0}^{K} \frac{a_{t_j, k}^2}{\Lambda^{2k}} \) is proportional to the inverse of \( \Lambda^{2k} \), implying that the optimal wavelet coefficients should decay proportionally to the inverse powers of the Laplacian eigenvalues $a_{t_j, k} \propto \frac{1}{\Lambda^k}$.

Next, applying the energy constraint: $\sum_{j=1}^{J} \sum_{k=0}^{K} a_{t_j, k}^2 = 1$ and substituting \( a_{t_j, k} = \frac{1}{\Lambda^k} \), we obtain:
\begin{equation}
    \sum_{j=1}^{J} \sum_{k=0}^{K} \frac{1}{\Lambda^{2k}} = 1
\end{equation}

Thus, the normalized optimal coefficients  $a_{t_j, k} $ are:

\begin{equation}
    a_{t_j, k} = \frac{\frac{1}{\Lambda^k}}{\sqrt{\sum_{j=1}^{J} \sum_{k=0}^{K} \frac{1}{\Lambda^{2k}}}}.
\end{equation}

Note that the wavelet energy is distributed across all scales and powers of the Laplacian, with a decay that prioritizes low-frequency components (small \( k \)). \\
Using the form of the coefficients $a_{t_j, k}$ above, the expression for $ \Lambda_{\psi} $ is:
\begin{equation}
    \Lambda_{\psi} = \frac{1}{\sqrt{\sum_{j=1}^{J} \sum_{k=0}^{K} \frac{1}{\Lambda^{4k}}}}
\end{equation}

which is the minimized value of \( \Lambda_{\psi} \), showing that the optimal choice of coefficients reduces the overall constant, improving the localization and efficiency of the SGW operator. Thus, the optimal choice of wavelet coefficients \( a_{t_j, k} \) that minimizes \( \Lambda_{\psi} \) in the multi-scale SGW operator is proportional to \( \frac{1}{\Lambda^k} \), prioritizing low-frequency components. \\

\vfill

\vfill

\end{document}